\documentclass[10pt,twocolumn,letterpaper]{article}

\usepackage{cvpr}              

\usepackage[table]{xcolor}
\definecolor{uoftcoolgray}{RGB}{163, 169, 173}

\renewcommand{\paragraph}[1]{\vspace{.5em}\noindent\textbf{#1.}}

\usepackage{booktabs}   
\usepackage{array}      
\usepackage{graphicx} 
\usepackage[accsupp]{axessibility}  
\usepackage{makecell}  
\usepackage[all]{nowidow}
\usepackage{siunitx}   
\usepackage{caption} 
\usepackage{float}
\usepackage[table]{xcolor}  
\usepackage{xcolor}
\definecolor{cvprblue}{rgb}{0.21,0.49,0.74}

\definecolor{uoftblue}{RGB}{30, 55, 101} 

\definecolor{uoftsecondaryblue}{RGB}{0,127,163} 

\definecolor{uoftpurple}{RGB}{109,36,122} 
\definecolor{uoftwarmred}{RGB}{220,70,51} 
\definecolor{uoftcoolblue}{RGB}{111,199,234} 
\definecolor{uoftteal}{RGB}{0,161,137} 
\definecolor{uoftfuchsia}{RGB}{171,19,104} 
\definecolor{uoftdarkgreen}{RGB}{13,83,77} 
\definecolor{uoftyellow}{RGB}{241,197,0} 
\definecolor{uoftlightgreen}{RGB}{141,191,46} 
\definecolor{staicrs}{RGB}{74,144,226} 
\definecolor{fencce}{RGB}{39,94,66}  
\definecolor{uoftcoolgray}{RGB}{208,209,201} 

\usepackage[pagebackref,breaklinks,colorlinks,allcolors=uoftsecondaryblue]{hyperref}

\def\paperID{} 
\def\confName{CVPR}
\def\confYear{2026}

\title{Material Magic Wand: \\ Material-Aware Grouping of 3D Parts in Untextured Meshes}

\author{
Umangi Jain$^{1}$\thanks{Work done partially during internship at Adobe}
\quad
Vladimir Kim$^{2}$
\quad
Matheus Gadelha$^{2}$
\quad
Igor Gilitschenski$^{1}$\textsuperscript{\dag}
\quad
Zhiqin Chen$^{2}$\textsuperscript{\dag} \\
$^{1}$University of Toronto, $^{2}$Adobe Research
}

\begin{document}
\maketitle

\begingroup
\renewcommand\thefootnote{\dag}
\footnotetext{Equal advising}
\endgroup

\begin{figure*}[t]
  \centering
   \includegraphics[width=0.9\linewidth]{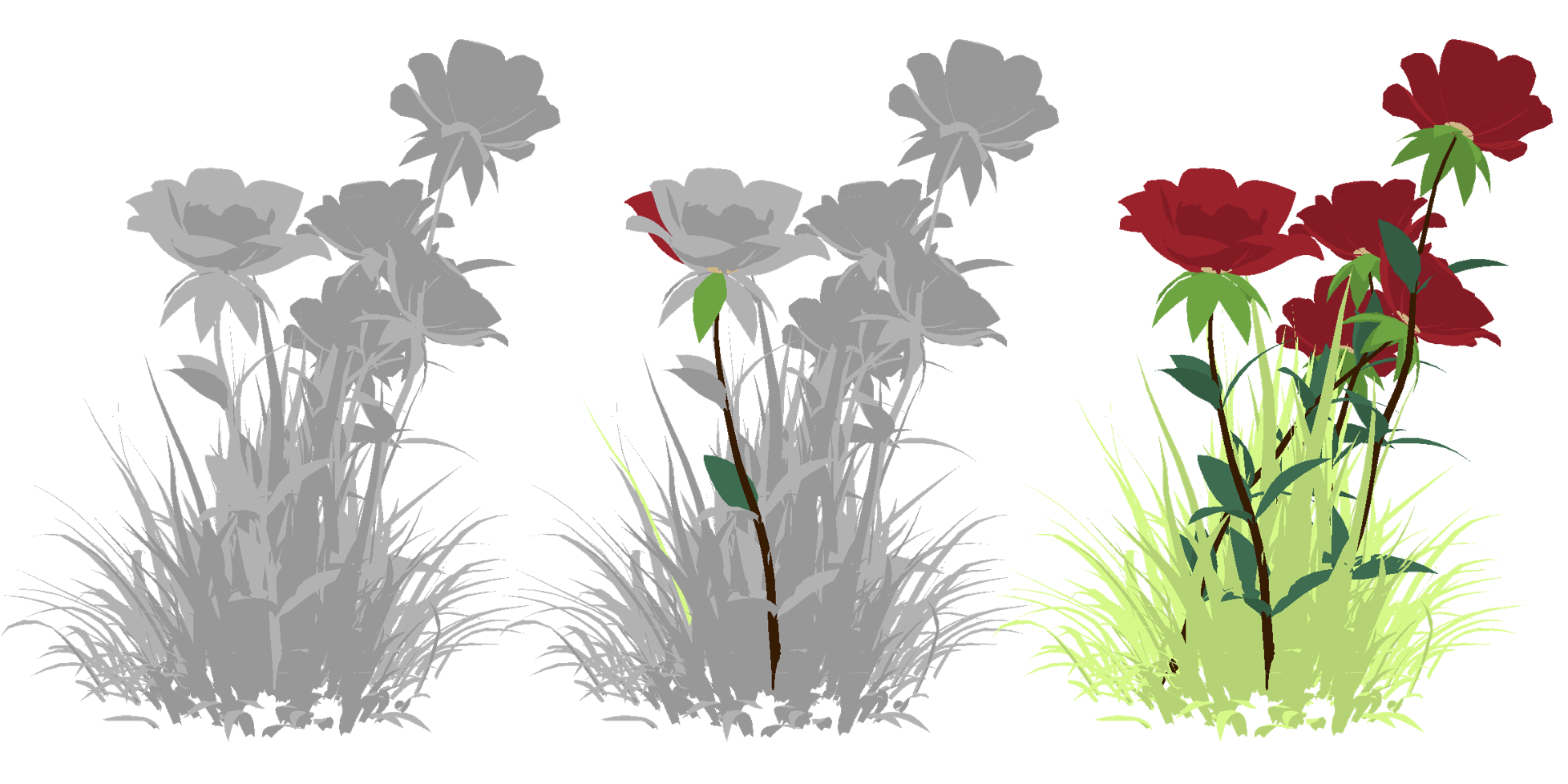}
   \caption{\textbf{Material Magic Wand.} Given an untextured 3D mesh (left) with existing part segmentation, which is often obtained by finding connected components of the mesh, a user can apply our tool to select a group of material-consistent parts by clicking on one single representative part. In the middle, we show example selections of \textcolor[HTML]{a60028}{petal},  
   \textcolor[HTML]{D3B683}{base}, \textcolor[HTML]{5DA836}{sepal},  \textcolor[HTML]{3b1a01}{stem}, \textcolor[HTML]{2E6E4E}{leaf}, and \textcolor[HTML]{BBD280}{grass}.
   For each selection, the tool automatically finds all other parts in the shape that are likely to share the same material (right) through geometric and contextual cues, accelerating the material assignment process. Colors may appear darker due to backface shading.}
   \label{fig:introduction}
\end{figure*}

\begin{abstract}
We introduce the problem of material-aware part grouping in untextured meshes.
Many real-world shapes, such as scales of pinecones or windows of buildings, contain repeated structures that share the same material but exhibit geometric variations.
When assigning materials to such meshes, these repeated parts often require piece-by-piece manual identification and selection, which is tedious and time-consuming.
To address this, we propose Material Magic Wand, a tool that allows artists to select part groups based on their estimated material properties -- when one part is selected, our algorithm automatically retrieves all other parts likely to share the same material. 
The key component of our approach is a part encoder that generates a material-aware embedding for each 3D part, accounting for both local geometry and global context.
We train our model with a supervised contrastive loss that brings embeddings of material-consistent parts closer while separating those of different materials;
therefore, part grouping can be achieved by retrieving embeddings that are close to the embedding of the selected part.
To benchmark this task, we introduce a curated dataset of 100 shapes with 241 part-level queries.
We verify the effectiveness of our method through extensive experiments and demonstrate its practical value in an interactive material assignment application. Project Page: \url{https://umangi-jain.github.io/material-magic-wand}.

\end{abstract}

\section{Introduction}
\label{sec:intro}

Many 3D shapes contain groups of related parts that share a common structural form while varying in their precise geometry. 
Such families of parts arise naturally in both natural and man-made objects. 
For instance, a pinecone may contain hundreds of individual scales that share a common form but differ in shape, orientation, and size; similarly, windows in buildings and vehicles are often repeated but non-identical.
In a 3D modeling scenario, such repeated parts often share the same material.
Yet in current modeling tools, repeated elements must be selected and assigned materials individually.
This makes material assignment repetitive and time-consuming, and its cost grows with the number of repeated elements and the mesh complexity.

To address this, we introduce the task of \emph{material-aware part grouping}: it assumes an \emph{untextured} 3D mesh already decomposed into fine-grained parts and a query part as input. The goal is to retrieve other parts within the shape that are likely to share the same material.
Despite the importance of material assignment in asset creation, the problem of grouping material-consistent parts under geometric variation remains largely unaddressed.

Existing works tackle related but fundamentally different problems. 
3D part segmentation methods~\cite{liu2025partfield, abdelreheem2023satr} aim to partition geometry into semantically meaningful components, whereas our task is higher-level grouping, not initial decomposition.
In our setting, the fine-grained mesh parts are already given, because in a typical 3D modeling workflow, an artist-created mesh can be naturally decomposed into parts defined by connected components, which we use as the smallest elements to be grouped.
Shape retrieval methods~\cite{chen2003visual, liu2023openshape} focus on comparing whole shapes across the database, rather than identifying related parts within a single shape. 
Symmetry detection methods~\cite{pauly2008discovering} generally rely on exact or near-isometric correspondences, and thus do not capture the breadth of structural similarity that arises in repeated but non-identical elements. 
Material segmentation methods~\cite{li2024materialseg3d,fischer2024sama} rely on textures to provide hints on material properties, and achieve material segmentation by adopting foundation image segmentation models, therefore facing challenges when handling small parts due to resolution limits.
To our knowledge, no existing method or benchmark evaluates grouping pre-segmented parts based on material consistency.

In this work, we present \emph{Material Magic Wand}.
Just like the Magic Wand tool in Photoshop, where a user can easily select pixels of similar colors by clicking on one pixel, Material Magic Wand enables artists to select parts that share the same material by clicking on one part, see Figure~\ref{fig:introduction}.
And similar to Magic Wand's \emph{Tolerance} parameter to balance between selecting more pixels with less similar colors or fewer pixels with more similar colors, Material Magic Wand can select more parts with less confidence or fewer parts with more confidence by tuning a threshold parameter, enabling fine-grained control and hierarchical selection, see Figure~\ref{fig:hierarchical_retrieval}.
Our tool can significantly speed up the material assignment process, allowing designers to assign materials to hundreds of parts with a single interaction.

Our key insight is to embed each part into an embedding space that encodes material similarity, so that part grouping can be done by retrieving embeddings that are close to the embedding of the query part.
Deriving such an embedding space is nontrivial, as both classic geometric descriptors~\cite{chen2003visual} and latest image embedding models such as DINO~\cite{oquabdinov2} or SigLIP~\cite{tschannen2025siglip} lack material understanding.
We therefore design a part encoder model that learns material-aware embeddings from large-scale collections of 3D shapes. 
Our encoder generates an embedding code for each 3D part, taking as input multiple images of the part, which are rendered in specific configurations to capture both local geometry and global context.
We train the encoder with a supervised contrastive loss to place the embeddings of parts that share the same material close together in the embedding space, while separating the embeddings of parts that differ in their material identity.

To our knowledge, there is no existing 3D dataset for material-aware grouping.
However, from the 3D shapes available in Objaverse~\cite{objaverse}, we are able to curate a dataset of approximately 1.9 million parts across 22,000 meshes with reliable material annotations to supervise our training.
Nonetheless, ambiguities naturally arise in material-aware grouping, so that parts can be grouped in various ways under different interpretations of the scene (see Figure~\ref{fig:inconsistent_materials}).
To reduce the noise caused by such ambiguities in evaluation, we propose a benchmark of 100 shapes and define 241 part groups for retrieval.
Shapes in the benchmark feature repeated but geometrically varied structures, and we manually refine their part-material associations to create clean ground truth for qualitative and quantitative studies.
Experiments are conducted on the benchmark, comparing our approach against various baselines, confirming its effectiveness.
In addition, we demonstrate Material Magic Wand in an interactive material assignment application, which further exemplifies the appealing capabilities offered by our method.

In summary, our main contributions are:
\begin{itemize}
\item We introduce the task of material-aware part grouping on untextured and pre-segmented meshes.
\item We curate a sizable 3D dataset for material-aware grouping and a benchmark for evaluation.
\item We propose Material Magic Wand to address the material-aware part grouping. Our method enables artists to interactively select material-consistent part groups efficiently. 
\end{itemize}

\section{Related Work}\label{relatedwork}
We situate our work in relation to prior research in part segmentation, shape retrieval, symmetry and repetition analysis, and material prediction.

\vspace{0.5\baselineskip}

\noindent\textbf{Part segmentation} aims to decompose a 3D shape into semantically meaningful subparts. 
The introduction of large-scale part segmentation datasets~\cite{yi2016scalable,mo2019partnet} enabled supervised training of deep neural networks and marked a major step toward scalable part segmentation. 
More recently, advances in 2D vision foundation models, such as Segment Anything (SAM)~\cite{kirillov2023segment,ravisam}, CLIP~\cite{radford2021learning}, and GLIP~\cite{li2022grounded}, have catalyzed the development of open-world 3D segmentation models.
Several methods propagate 2D SAM masks to 3D for shape decomposition~\cite{tang2024segment, kim2024garfield, ying2024omniseg3d, he2024view}, or train feedforward networks that segment parts using 3D point prompts~\cite{lang2024iseg,liu2025partfield,zhou2024point}.
Others~\cite{abdelreheem2023satr, kim2024partstad, liu2023partslip, abdelreheem2023zero, xue2025zerops, zhong2024meshsegmenter, zhou2023partslippp, ma2025find, umam2024partdistill} achieve text-based part segmentation by applying open-world 2D detectors on rendered multi-view images and fusing the predictions.
Distinct from part segmentation and component labeling methods~\cite{wang2018learning,jones2022neurally}, our work assumes that parts are already segmented into their finest level, and we focus on grouping the parts based on their potential material assignment.

\vspace{0.5\baselineskip}

\noindent\textbf{Shape matching and retrieval} aim to retrieve 3D shapes from a database that are similar to a query shape. The query could be sketch-based~\cite{wang2015sketch}, image-based~\cite{wu2024generalizing}, text-based~\cite{ruan2024tricolo}, and 3D shape-based. 
For 3D shape-based retrieval, global shape descriptors, both classic ones~\cite{ankerst19993d, hilaga2001topology, chen2003visual} and deep learning-based~\cite{xie2015deepshape, fang20153d, liu2023openshape, wang2025describe}, compare 3D shapes at the object level.
These approaches seek inter-shape geometric similarity, while our work targets intra-shape semantic similarity: we retrieve similar parts within a single shape based on their estimated material properties.

\vspace{0.5\baselineskip}

\noindent\textbf{Symmetry and structural repetition} methods detect symmetry, periodicity, and near-regular structure in 3D geometry.
Early work focuses on global or local symmetry detection using geometric hashing or feature correspondences~\cite{podolak2006planar,mitra2006partial}. 
Subsequent research considers partial and approximate symmetries and near-regular structures~\cite{mitra2013symmetry,lipman2010symmetry}, as well as repeated elements and their spatial organization~\cite{pauly2008discovering,huang2014near}. 
These approaches generally assume repeated elements to be related by rigid or near-isometric transformations, and work best when the intra-group variation is relatively small.
In contrast, the parts we aim to group often vary significantly in shape and proportion.

\vspace{0.5\baselineskip}

\noindent\textbf{Material segmentation and assignment}
 have been extensively studied in the image domain~\cite{sharma2023materialistic, guerrero2025fine}.
In 3D, classical methods~\cite{chajdas2010assisted} define hand-crafted rules to compute surface similarity for matching textured surfaces.
Recent approaches~\cite{li2024materialseg3d, fischer2024sama} perform material-aware segmentation by finetuning SAM and fusing multi-view predictions, but the view resolution in these methods often limits their ability to capture small parts.
They also assume textured meshes, where colors provide cues about material properties.
For untextured meshes, most approaches~\cite{huang2025material,zhang2024dreammat} generate material maps for whole objects, inheriting the same resolution issues and offering limited editability due to the lack of material segmentation.

\begin{figure*}[h]
\vspace{-20pt}
  \centering
   \includegraphics[width=1.0\linewidth]{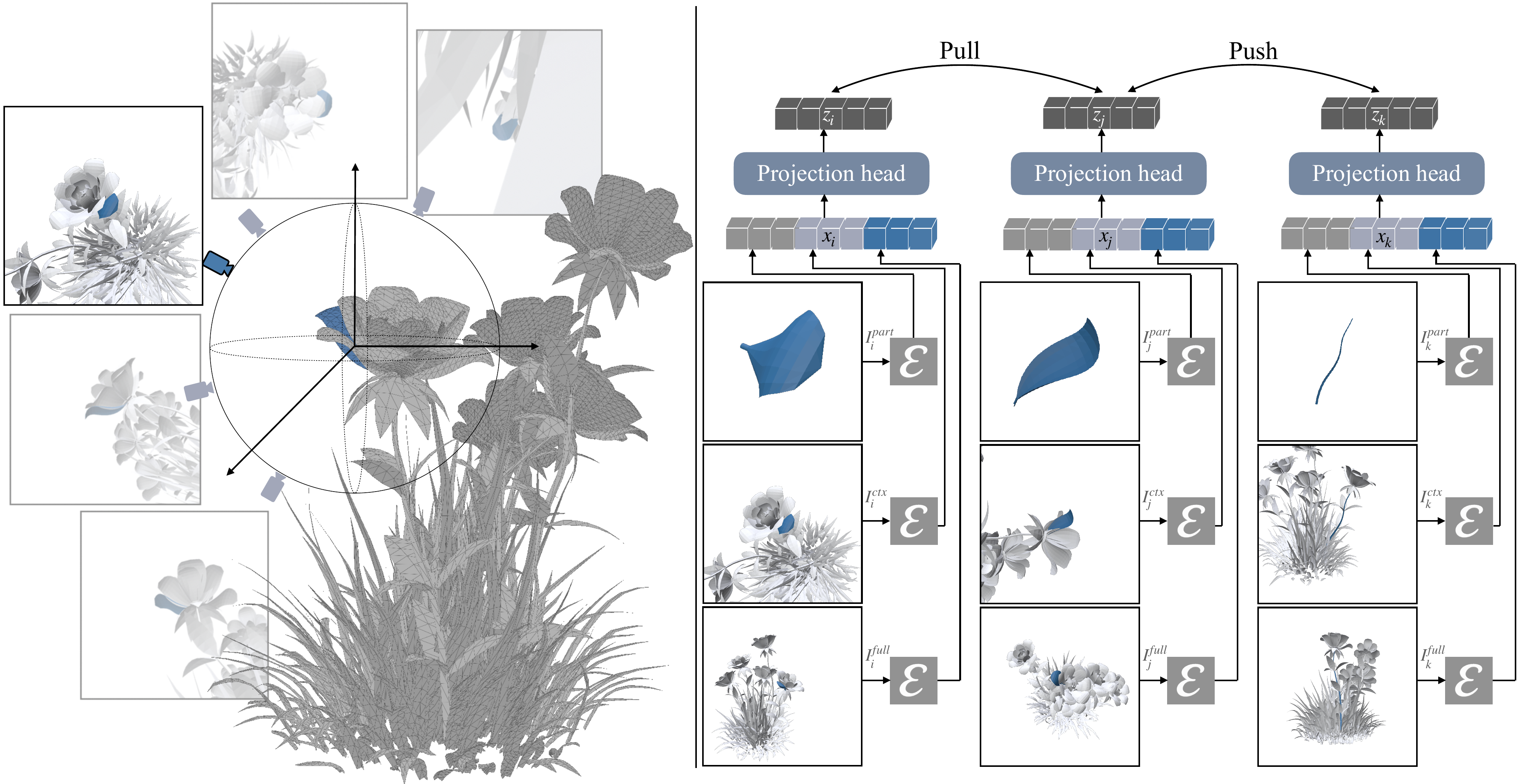}
    \caption{\textbf{Method Overview.} \textit{Left:} Our view selection process renders each part with nearby context from multiple viewpoints sampled randomly over a hemisphere. We choose the one with minimal occlusion, $I^{ctx}$, and use the same viewpoint to render the part in isolation, $I^{part}$. $I^{full}$ captures the entire mesh. We highlight the part with a different color from the rest of the mesh. \textit{Right:} For each part, its corresponding images are passed through an encoder and their embeddings are concatenated. During training, embeddings of parts with the same material are pulled together, while those with different materials are pushed apart.}
   \label{fig:method}
\end{figure*}

\section{Method}

Given an untextured 3D mesh with existing part segmentation, which is often obtained by finding connected components of the mesh, our method learns a material-aware embedding space for its parts, where each part is encoded by our encoder network through rendered views.

\paragraph{Notations}
Assuming a mesh $\mathcal{S}$ is segmented into parts $p_i$, and associated with a material label $y_i$.
We aim to learn a part encoder network that maps each part $p_i$ into a latent embedding $z_i$, such that ideally, $z_i=z_j$ if and only if $y_i=y_j$.
For part $p_i$, we define its positive set as $P_i = \{j | j \neq i, y_j = y_i\}$, which contains all other parts in the mesh $\mathcal{S}$ that share the same material with $p_i$;
and its complement set as $A_i = \{j | j \neq i\}$, containing all parts in the mesh except $p_i$.
To encode the parts, each part $p_i$ is represented using three rendered images, defined as $ [I_i^{\text{part}}, I_i^{\text{ctx}}, I_i^{\text{full}}]$, corresponding to the \textit{isolated-part} view, \textit{part-with-context} view, and \textit{full-object} view, which will be explained later.

\paragraph{Part Encoder}
We adopt a foundation vision model $\mathcal{E}$ to encode the three rendered images individually, initializing the backbone with the DINO-v3 \textit{small} model~\cite{simeoni2025dinov3} and finetuning the last three transformer blocks.
The features from all three images are concatenated to obtain an 1152-d embedding vector $x_i$, and we use a projection head $f$ to map the features into the contrastive latent space to obtain an 128-d embedding $z_i$, see Figure~\ref{fig:method}.
The projection head is a two-layer multilayer perceptron (MLP) with ReLU~\cite{nair2010rectified} activation.
The resulting embedding $z_i$ is $\ell_2$-normalized.
$$
x_i = \big[\, \mathcal{E}(I_i^{\text{part}});  \mathcal{E}(I_i^{\text{ctx}});  \mathcal{E}(I_i^{\text{full}}) \,\big]
; \;
z_i = \frac{f(x_i)}{\|f(x_i)\|_2}.
$$
\paragraph{Training objective}
We would like for parts with the same material to have nearby embeddings, and for parts with different materials to be far apart.
Therefore, we adopt the Supervised Contrastive Loss~\cite{khosla2020supervised} for learning the embedding space.
Our objective is:
$$
\mathcal{L}
=
\underset{\mathcal{S}}{\mathbb{E}} \; \underset{i}{\mathbb{E}} \underset{j \in P_i}{\mathbb{E}}
\left(
  -\log
  \frac{
    \exp(z_i \cdot z_j / \tau)
  }{
    \displaystyle\sum_{a \in A_i}
    \exp(z_i \cdot z_a / \tau)
  }
\right),
$$
where $\tau$ is the temperature parameter and $\cdot$ is dot product.
In practice, we put all parts in a training batch except for $p_i$ itself to $A_i$ to stablize training.




\paragraph{Inference}
We find that instead of using the compressed embedding $z_i$, using the original higher-dimensional embedding $x_i$ leads to consistently better performance (see Table \ref{tab:ablation}).
Therefore, we compute embedding $x_i$ for each part $p_i$ in the 3D mesh.
We define the similarity metric between two parts $p_i$ and $p_j$ as the negative $\ell_1$ distance between their embeddings: $s(p_i, p_j) = -\|x_i - x_j\|_1$.
Then, given a query part $p_i$, we can select a group of parts $\{ p_j | s(p_i, p_j) \leq \lambda \}$, where a higher threshold~$\lambda$ includes more parts, providing a looser grouping, and a lower~$\lambda$ restricts the selection to the most similar parts.



\paragraph{View selection}
To encode part-level representations, we render each part under three contexts: \textit{isolated-part}, \textit{part-with-context}, and \textit{full-object}, see Figure~\ref{fig:method}.
In the isolated part view $I_i^{\text{part}}$, only the part $p_i$ is shown in a highlighted color and other parts are hidden.
In the part-with-context view $I_i^{\text{ctx}}$, we render part $p_i$ as well as other parts of the mesh, while only $p_i$ has highlighted color. We adjust the camera distance, so that in the rendered view, $p_i$ occupies around 25\% of space in one image dimension.
The full-object view $I_i^{\text{full}}$ renders the entire mesh, while $p_i$ is also highlighted.
For $I_i^{\text{ctx}}$, we render multiple views from different viewpoints, and select the best view following the perceptual preference for maximizing visible surface area~\cite{secord2011perceptual}.
Specifically, we sample 16 candidate camera positions on a hemisphere surrounding the part and choose the one view that maximizes the visible area of the part in the rendered image.
If the part is heavily occluded under all candidate views, we reduce the context region by zooming the camera towards the part.
For $I_i^{\text{part}}$, we use the same viewpoint as $I_i^{\text{ctx}}$ but with camera zoomed-in on the part.
For $I_i^{\text{full}}$, we place the camera along the direction from the object center to the part centroid.
Our method targets untextured meshes, so we remove all material and texture before rendering.

\paragraph{Part deduplication}
Meshes can have duplicated parts - parts that are essentially identical but have undergone transformations resulting in different sizes and orientations.
Identifying duplicated parts from rigid transformations is relatively easy with a histogram matching algorithm~\cite{ankerst19993d, Thea}.
Therefore, we run the algorithm to group identical parts in each mesh, and randomly select one exemplar part in each group to compute the embedding for all the parts in the group.
This deduplication step reduces computational cost.

\paragraph{Training dataset}
Recent datasets such as Material3D~\cite{huang2025material} and DreamMat~\cite{zhang2024dreammat} provide textured 3D meshes for training material prediction and generation models.
However, these datasets assign materials at the surface level, not part level, therefore they cannot be directly used for part-level grouping.
Our task requires a large-scale 3D dataset consisting of shapes with per-part material ID, where the same material ID is shared by different parts.
Therefore, we curate a subset of 22,000 meshes with material assignment from Objaverse~\cite{objaverse}. 
Since Objaverse does not provide fine-grained part segmentation, we extract parts using connected components, after performing vertex merging on the mesh to avoid fragmented components.
We assign each part a material ID by taking the majority material label over the faces it contains.
Material IDs are defined independently for each mesh.
Meshes in Objaverse also present several challenges, especially the imbalanced material distribution, e.g., 99\% of the materials are only used once on one part, or one material is used on 99\% of the parts.
We apply data balancing strategies to mitigate both within and across mesh imbalance, resulting in a more uniform distribution over both meshes and material IDs. See appendix for more details.

\paragraph{Implementation details}
We use an OpenGL-based renderer for generating the training data at scale. 
All images are rendered at 512 $\times$ 512 resolution. 
We train the model using the Adam optimizer~\cite{kingma2014adam} with learning rate $1\times10^{-5}$.
The model is trained for 20,000 steps with a batch size of 256.
We observe marginal difference when scaling to a larger encoder backbone; see ablation study in Table~\ref{tab:ablation}.

\begin{table*}[t]
\rowcolors{2}{white}{uoftcoolgray!25}
    \centering
    \caption{Comparison of geometry-based, vision-embedding, and segmentation-based baselines on our material-aware grouping task. All metrics are macro-averaged and reported in percentage. Our method achieves consistent improvements across all evaluation metrics.}
    \label{tab:main_results}
    \setlength{\tabcolsep}{5pt}
    \renewcommand{\arraystretch}{1.1}
    \begin{tabular}{lcccccccc}
        \toprule
        \textbf{Method} & \textbf{AUC PR} & \textbf{R-Prec} & \textbf{mAP} & \multicolumn{4}{c}{\textbf{Recall@K}} & \textbf{F1 }\\
        \cmidrule(lr){5-8}
         &  &  &  & \textbf{R@5} & \textbf{R@10} & \textbf{R@20} & \textbf{R@100} \\
        \midrule
        \hspace{1em}Histogram Matching~\cite{Thea} & 26.85 & 25.45  & 30.71 & 4.55 & 8.89 & 15.97 & 45.64  & 23.84\\
        \hspace{1em}SigLIP-v2~\cite{tschannen2025siglip}               & 62.83 &  56.02 & 60.58 & 22.88 & 31.20 & 40.72 & 68.44 &  39.44   \\
        \hspace{1em}PartField~\cite{liu2025partfield}          & 75.30  & 67.74 & 70.52 & 26.88 & 37.12 & 47.92 &  71.31 & 56.57 \\
        \hspace{1em}DINO-v3 $\textit{small}^\dagger$~\cite{simeoni2025dinov3}       &   \underline{81.14} & \underline{78.32}  & \underline{83.49} & \underline{34.57} & \underline{45.49} & \underline{56.63} & \underline{82.18} &  \underline{59.36}  \\
        
        \hspace{1em}\textbf{Ours}               &   \textbf{89.74} & \textbf{88.33} & \textbf{91.70} & \textbf{37.99} & \textbf{50.98} & \textbf{62.79} & \textbf{83.67} & \textbf{75.94}  \\
        \bottomrule
        \multicolumn{9}{c}{\footnotesize \textsuperscript{$\dagger$}We evaluate several DINO variants (v2/v3 at different scales) and report the one achieving the highest AUC.}
    \end{tabular}
\end{table*}

\section{Experiments}

\begin{figure}[t]
\vspace{-20pt}
  \centering
   \includegraphics[width=\linewidth]{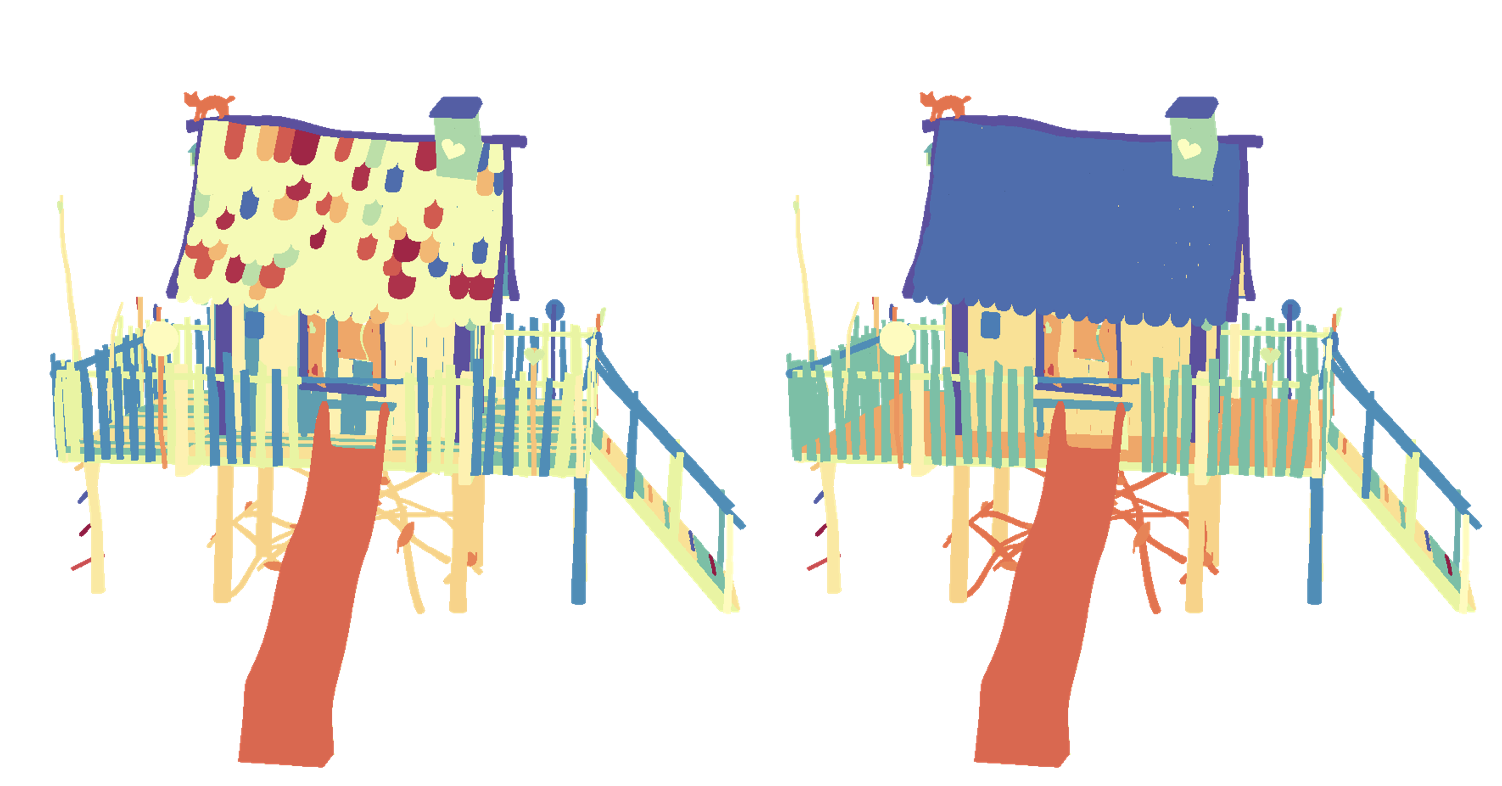}

   \caption{\textbf{Ambiguities in raw material IDs.} Using meshes from Objaverse directly for material grouping can introduce ambiguity in the testing benchmark due to artistic intent, noisy labeling, or coarse material assignment. For example, some shingles on the roof exhibit mixed materials, and fence or wall slats alternate between different material labels (left). To reduce such inconsistencies, we manually refine the material annotations to make the material assignment more uniform and fine-grained (right).}
   \label{fig:inconsistent_materials}
\end{figure}

\paragraph{Evaluation Benchmark}
Existing datasets do not provide part-level groupings within a single mesh that reflect fine-grained material-level similarity.
Therefore, we construct a testing benchmark dataset that features repeated but geometrically varied structures in each shape.
The material assignments in Objaverse can exhibit inconsistencies: materials may be shared across parts that differ functionally and in appearance, while in some cases, artistic intent can introduce additional ambiguity by assigning different materials to the same type of parts.
For instance, in Figure~\ref{fig:inconsistent_materials}, roof shingles and fence slats have a variety of material types. 
Such cases introduce ambiguity that would lead to unreliable evaluation.
To obtain consistent ground-truth, we manually refine the material assignments for 100 meshes from Objaverse in Blender, thereby resolving ambiguous cases.
This results in 241 query parts with well-defined ground-truth retrieval sets, forming our benchmark for both qualitative and quantitative evaluation.

The benchmark spans a wide range of mesh complexity.
Across the 100 meshes, the number of parts per mesh varies substantially (median = 265; IQR = 1{,}144; range = 16–40{,}086), reflecting large differences in granularity. 
The 241 ground-truth part groups also exhibit significant variations in size (median 20; IQR = 72; range = 2-32{,}267).
For evaluation, an arbitrarily chosen part is designated as the query for the group, and all other parts in the same group are considered equally relevant positive matches.

\paragraph{Metrics}
For each query, we evaluate retrieval as a ranked list over all other parts within the same mesh. 
We report standard retrieval metrics: area under the precision–recall curve (AUC PR),  R-Precision (R-Prec), mean average precision (mAP), and Recall@k. 
All metrics are reported using macro-averaging (each query weighted equally) to prevent queries with large ground-truth sets from dominating the scores.
To assess grouping performance, we report the average F1 score, computed between the predicted and ground-truth groups for a given similarity threshold~$\lambda$.
The threshold~$\lambda$ is selected separately for each method to ensure fairness, using the value that maximizes F1 on a small held-out validation split (5 meshes, 13 queries).

\begin{figure}
\vspace{-20pt}
  \centering
   \includegraphics{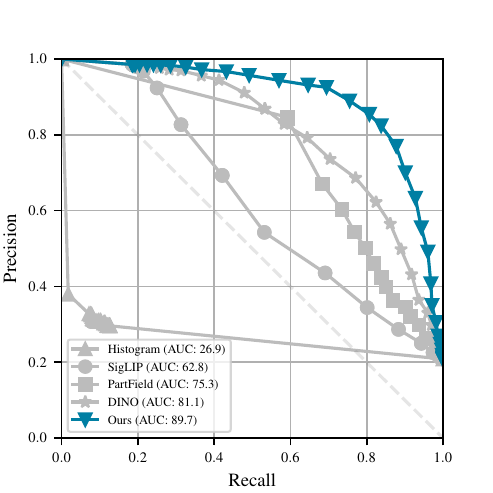}

   \caption{\textbf{Precision–Recall curve.} We sweep the similarity threshold to evaluate retrieval performance. Our method consistently maintains higher precision across all recall levels, achieving the highest AUC (89.7), followed by DINO (81.1).}
   \label{fig:auc}
\end{figure}

\begin{figure*}
  \centering
   \includegraphics[width=0.95\linewidth]{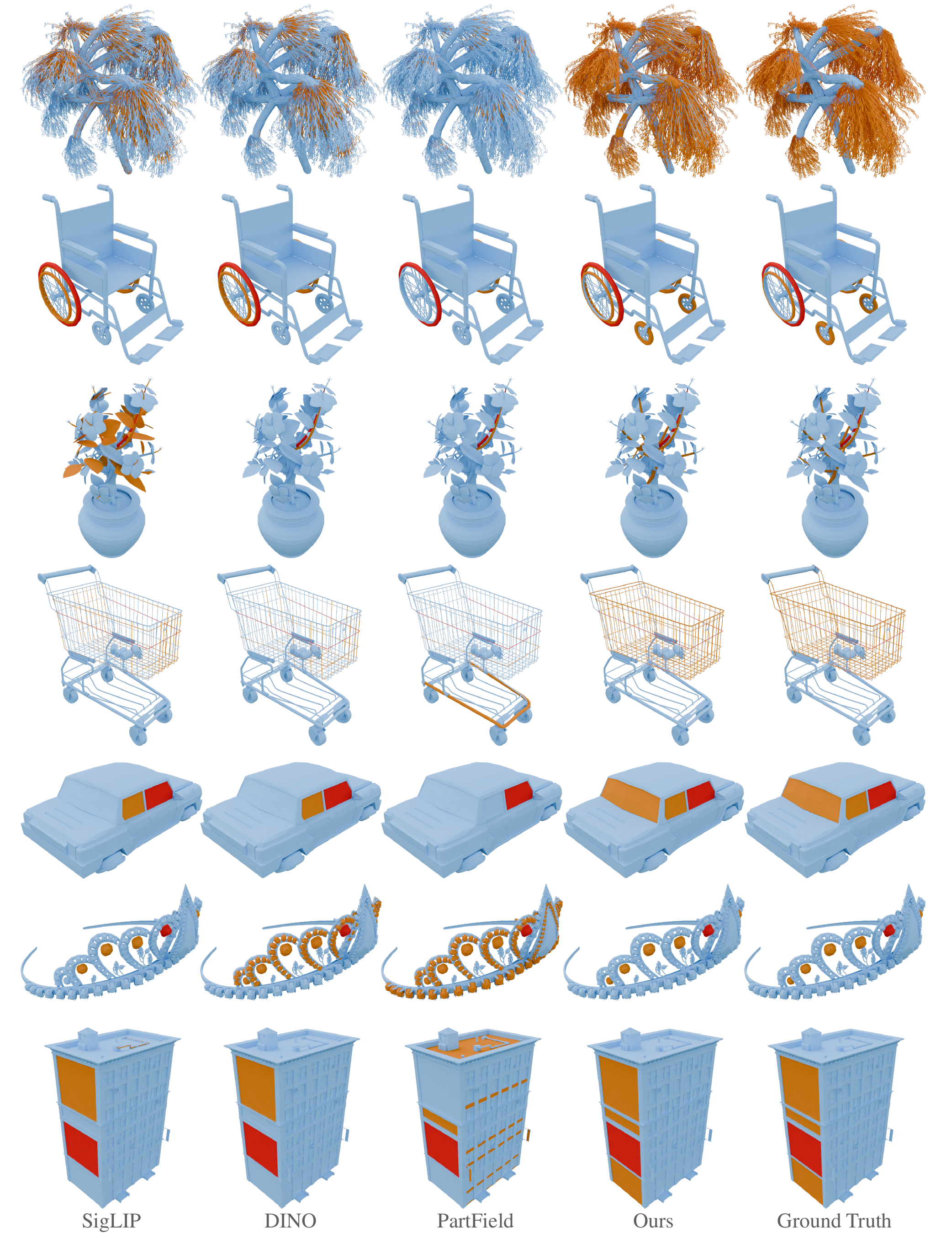}
   \caption{\textbf{Qualitative comparison.} For each mesh, the red part denotes the query, and orange parts indicate the retrieved matches. Our method retrieves components that are both geometrically and contextually similar with the query, while baselines often miss structurally related parts or include visually similar but contextually incorrect ones.}
   \label{fig:qual_results}
\end{figure*}

\begin{figure*}
  \centering
   \includegraphics[width=\linewidth]{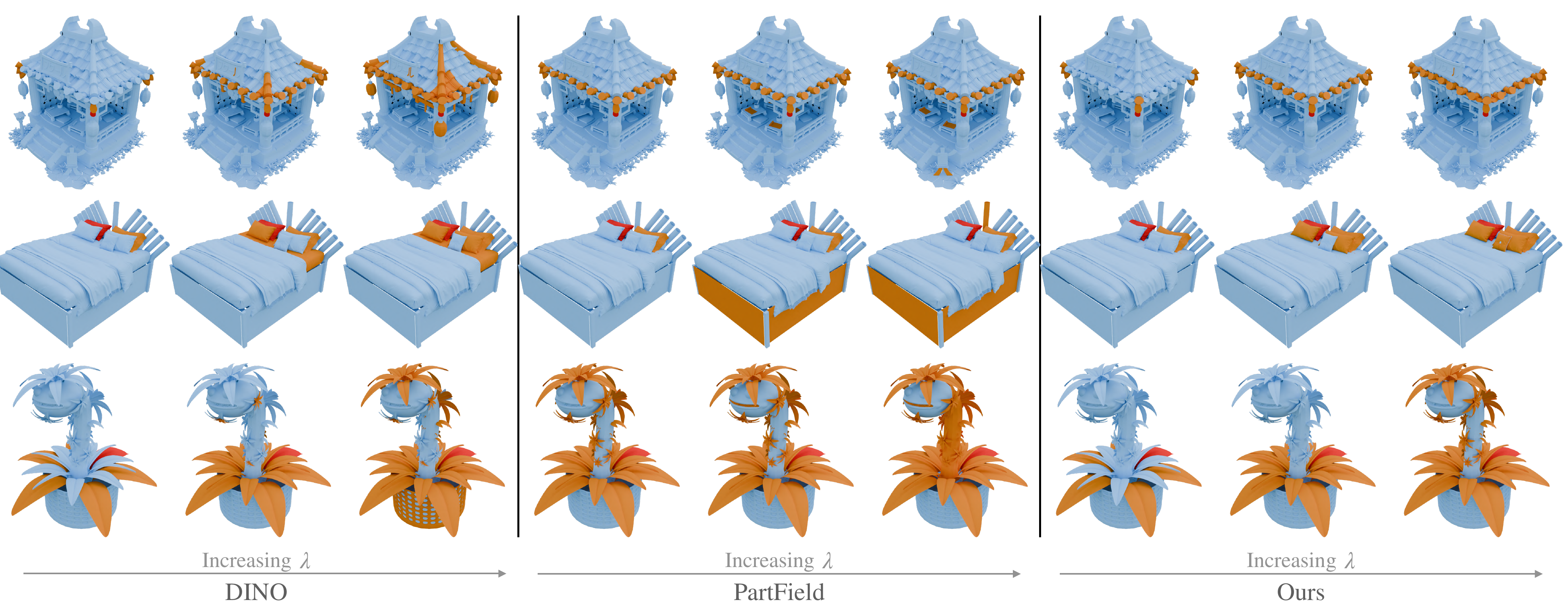}

   \caption{\textbf{Effect of changing the grouping tolerance parameter.} The similarity threshold $\lambda$ controls the cut-off distance in the parts' embedding space for part retrieval. With our method (right), at low values of $\lambda$, only highly similar parts are selected; increasing $\lambda$ gradually expands the selection to include less similar, yet related parts. In contrast, baselines exhibit less stable behavior by either retrieving unrelated components at higher thresholds or lacking fine-grained control at lower thresholds.}
   \label{fig:hierarchical_retrieval}
\end{figure*}

\paragraph{Comparison baselines}
We compare against three categories of baselines. 
(\textit{i}) \textbf{Histogram matching}~\cite{Thea}: a non-learning geometry-based baseline that uses statistics of the given mesh as features; we also use this algorithm with low tolerance threshold for part deduplication. 
(\textit{ii}) \textbf{Vision foundation model embeddings}: we render each part (isolated, with context, and full-object) and extract embeddings using DINO-v2~\cite{oquabdinov2}, DINO-v3~\cite{simeoni2025dinov3}, and SigLIP~\cite{zhai2023sigmoid}; results are reported for the best-performing backbone (we find minimal variance across these models; see Appendix). 
(\textit{iii}) \textbf{PartField}~\cite{liu2025partfield}: a hierarchical 3D part segmentation model that produces part-level embeddings; we test the released model directly on our benchmark.
We apply part deduplication to all baselines for a fair comparison.

\paragraph{Quantitative results}
Table \ref{tab:main_results} compares our method against geometric, vision foundation, and part-feature baselines across retrieval and grouping metrics. 
Our model achieves the highest performance on all measures, outperforming the strongest baseline (DINO-v3~\textit{small}) by a margin of $+8.6\%$ in AUC for retrieval and $+16.6\%$ in F1 score for grouping. 
Figure \ref{fig:auc} shows the precision-recall curve, where our method consistently maintains higher precision across the full recall range.
For computing the precision–recall curve, we sweep thresholds across the full range of similarity scores between the minimum and maximum values. To ensure uniform coverage under non-linear score distributions, we use quantile-based threshold sampling.
The geometric baseline (Histogram Matching) exhibits a steep precision drop as recall increases, indicating that purely shape-based descriptors are brittle and only effective for near-duplicate parts.
The weaker performance of PartField can be attributed to its different training objective: hierarchical part segmentation, which does not align with the goal of learning material-consistent embeddings.

\paragraph{Qualitative Results} Figure \ref{fig:qual_results} presents a visual comparison across methods. 
While DINO and SigLIP retrieve parts with similar visual appearance, they often miss structurally related components, for example, caster wheels in the wheelchair, vertical wire meshes in the shopping cart, or the rear windscreen in the car.
The embeddings from PartField, trained for hierarchical shape decomposition, tend to miss geometric correspondences. 
Our method incorporates contextual cues to retrieve material-consistent parts, such as the jewels on the crown that appear under the same spatial context, while avoiding visually similar but contextually different ones retrieved by DINO and PartField.

\paragraph{Effect of changing threshold} Material Magic Wand offers a grouping tolerance parameter that can be controlled via the similarity threshold $\lambda$.
Increasing $\lambda$ progressively expands the retrieved set from strictly matching parts to broader, structurally related regions.
As shown in Figure \ref{fig:hierarchical_retrieval}, our model exhibits smooth transitions: for instance, in the bed example, retrieval expands from the identical pillow to geometrically similar pillows to all pillows; likewise, in the pot example, retrieval grows from local leaves to the full set of leaves in the shape. In contrast, DINO and PartField start retrieving unrelated components as threshold increases. PartField also lacks fine-grained control, e.g., grouping most parts even at very low thresholds.

\begin{figure}
\vspace{-10pt}
  \centering
   \includegraphics[width=0.9\linewidth]{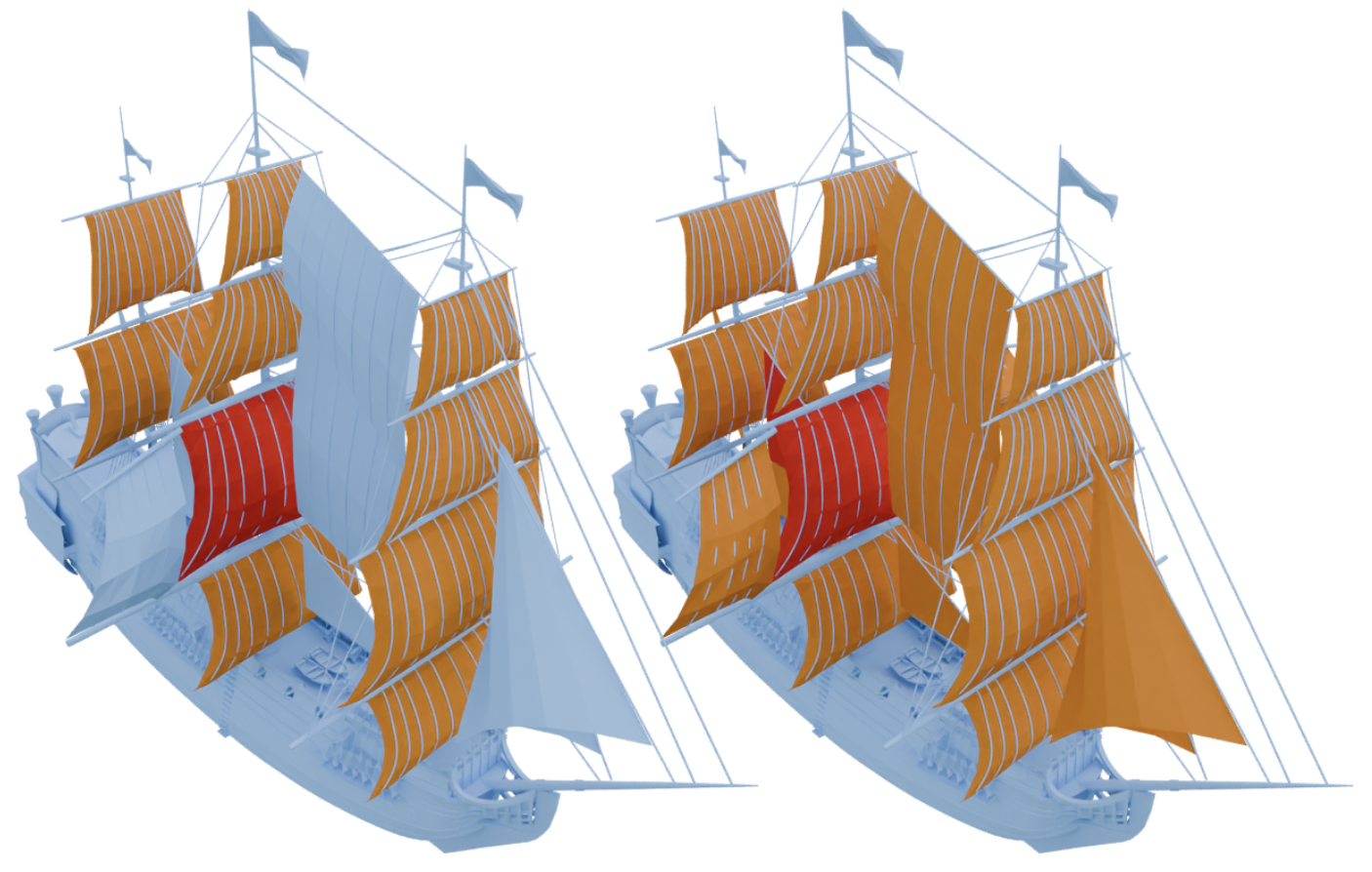}

   \caption{\textbf{Effect of multiple queries.} Additional query parts improve retrieval by capturing a more complete set of desired components. The initial query (left) retrieves only a subset of sails, while adding an additional example (right) helps recover the rest.}
   \label{fig:iterative}
    \vspace{-20pt} 
\end{figure}

\begin{table}
    \rowcolors{2}{white}{uoftcoolgray!25}
    \centering
    \caption{\textbf{Ablation results.} We examine the contribution of our design choices. Removing any component reduces performance, with the full model achieving the highest scores across all metrics.}
    \label{tab:ablation}
    \setlength{\tabcolsep}{5pt}
    \renewcommand{\arraystretch}{1.1}
    \begin{tabular}{lcccc}
        \toprule
        \textbf{Method} & \textbf{AUC} & \textbf{R-Prec} & \textbf{mAP} &  \textbf{R@20}\\
         w/o \textit{isolated part} & 86.90 & 84.97 & 88.53 & 61.11 \\
        w/o \textit{part-with-context} & 87.30 & 85.37 & 88.91 & 61.09\\
         w/o  \textit{full-object} & 88.89 & 87.72 & 90.75 & 62.36\\
         Only \textit{isolated part} & 86.18 & 81.88 & 85.88 & 59.87\\ \hline
         Init from Dino-v2 \textit{L} & 86.51 & 84.58 & 88.61 & 61.13 \\ 
         Random initialization & 78.54 & 74.52 & 79.62 & 56.25 \\
         Finetune last 5 blocks & 89.52 & 88.22 & 91.35 & 62.52 \\
          \hline
         Retrieval with $z$ & 87.45 & 87.58 & 90.90 & 62.34\\ \hline
         w/o data rebalancing & 78.41 & 76.22 & 80.39 & 55.83\\ \hline
         Ours & \textbf{89.74} & \textbf{88.33} & \textbf{91.70} & \textbf{62.79} \\
        \bottomrule
    \end{tabular}
\end{table}

\paragraph{Ablation study} We analyze the contribution from each design choice in Table \ref{tab:ablation}. 
Removing any of the three rendered views (\textit{isolated part}, \textit{part-with-context}, or \textit{full-object}) reduces performance, with the largest drop from removing the isolated part. 
To emulate a shape-retrieval-only setup, we also evaluate using only the isolated part rendering and observe drop in performance, highlighting the importance of contextual cues.
We test performance using pre-projection $x_i$ and post-projection embedding $z_i$ and find that the former performs better for retrieval.
Increasing the model size (to DINO large) or finetuning additional layers provides negligible gains, likely limited by the noisy supervision in Objaverse. 
Training from scratch or omitting data rebalancing sharply reduces performance, emphasizing the importance of strong initialization and balanced supervision (all variants are trained for same number of steps for fairness).

\paragraph{Multiple Clicks} While adjusting the tolerance provides control over the degree of similarity, part grouping can remain ambiguous depending on user intent or multiple valid choices. 
Our method can be applied iteratively, where providing additional query parts refines the retrieval by improving coverage of the desired components. 
As shown in Figure~\ref{fig:iterative}, the initial selection retrieves only geometrically similar sails, whereas adding another example also retrieves the jibs.
We show our tool in an interactive material assignment application in the supplementary video.

\paragraph{Limitations} Our method produces a deterministic ranking of parts, but grouping can involve multiple valid choices which we do not explicitly model.
In meshes with significant self-occlusion or many highly obstructed parts, our rendering-based view selection can struggle to produce clear contextual cues. See appendix for more details.


\section{Conclusion}\label{sec:conclusion}


We introduced Material Magic Wand, a framework for material-aware part grouping that learns to associate geometrically and contextually similar components within a mesh. 
Through contrastive training on large-scale data, our method enables robust retrieval and grouping, facilitating fast and more consistent material assignment (see Appendix for runtime). 
We also create a test benchmark for evaluating this task.
Our experiments demonstrate significant improvements over geometric and vision-based baselines, and qualitative results highlight its applicability to 3D modeling pipelines.

\clearpage
{
    \small
    \bibliographystyle{ieeenat_fullname}
    \bibliography{main}

@String(CVPR= {IEEE Conf. Comput. Vis. Pattern Recog.})

@String(ICCV= {Int. Conf. Comput. Vis.})

@String(ECCV= {Eur. Conf. Comput. Vis.})

@String(NIPS= {Adv. Neural Inform. Process. Syst.})

@String(TOG= {ACM Trans. Graph.})

@String(ICLR = {Int. Conf. Learn. Represent.})

@String(CVPR  = {CVPR})

@String(ICCV  = {ICCV})

@String(ECCV  = {ECCV})

@String(NIPS  = {NeurIPS})

@String(ICML  = {ICML})

@String(WACV  = {WACV})

@String(TOG   = {ACM TOG})

@String(ICLR  = {ICLR})

@article{tschannen2025siglip,
  title={Siglip 2: Multilingual vision-language encoders with improved semantic understanding, localization, and dense features},
  author={Tschannen, Michael and Gritsenko, Alexey and Wang, Xiao and Naeem, Muhammad Ferjad and Alabdulmohsin, Ibrahim and Parthasarathy, Nikhil and Evans, Talfan and Beyer, Lucas and Xia, Ye and Mustafa, Basil and others},
  journal={arXiv preprint arXiv:2502.14786},
  year={2025}
}

@inproceedings{mo2019partnet,
  title={Partnet: A large-scale benchmark for fine-grained and hierarchical part-level 3d object understanding},
  author={Mo, Kaichun and Zhu, Shilin and Chang, Angel X and Yi, Li and Tripathi, Subarna and Guibas, Leonidas J and Su, Hao},
  booktitle=CVPR,
  pages={909--918},
  year={2019}
}

@inproceedings{lang2024iseg,
  title={iseg: Interactive 3d segmentation via interactive attention},
  author={Lang, Itai and Xu, Fei and Decatur, Dale and Babu, Sudarshan and Hanocka, Rana},
  booktitle={SIGGRAPH Asia 2024 Conference Papers},
  pages={1--11},
  year={2024}
}

@inproceedings{zhou2024point,
  title={Point-sam: Promptable 3d segmentation model for point clouds},
  author={Zhou, Yuchen and Gu, Jiayuan and Chiang, Tung Yen and Xiang, Fanbo and Su, Hao},
  booktitle=ICLR,
  year={2025}
}

@inproceedings{liu2025partfield,
  title={Partfield: Learning 3d feature fields for part segmentation and beyond},
  author={Liu, Minghua and Uy, Mikaela Angelina and Xiang, Donglai and Su, Hao and Fidler, Sanja and Sharp, Nicholas and Gao, Jun},
  booktitle=ICCV,
  pages={9704--9715},
  year={2025}
}

@inproceedings{radford2021learning,
  title={Learning transferable visual models from natural language supervision},
  author={Radford, Alec and Kim, Jong Wook and Hallacy, Chris and Ramesh, Aditya and Goh, Gabriel and Agarwal, Sandhini and Sastry, Girish and Askell, Amanda and Mishkin, Pamela and Clark, Jack and others},
  booktitle=ICML,
  pages={8748--8763},
  year={2021},
  organization={PmLR}
}

@inproceedings{zhong2024meshsegmenter,
  title={Meshsegmenter: Zero-shot mesh semantic segmentation via texture synthesis},
  author={Zhong, Ziming and Xu, Yanyu and Li, Jing and Xu, Jiale and Li, Zhengxin and Yu, Chaohui and Gao, Shenghua},
  booktitle=ECCV,
  pages={182--199},
  year={2024},
  organization={Springer}
}

@inproceedings{li2024materialseg3d,
  title={Materialseg3d: Segmenting dense materials from 2d priors for 3d assets},
  author={Li, Zeyu and Gan, Ruitong and Luo, Chuanchen and Wang, Yuxi and Liu, Jiaheng and Zhu, Ziwei and Li, Qing and Yin, Xucheng and Zhang, Man and Zhang, Zhaoxiang and others},
  booktitle={Proceedings of the 32nd ACM International Conference on Multimedia},
  pages={370--379},
  year={2024}
}

@article{fischer2024sama,
  title={SAMa: Material-aware 3D selection and segmentation},
  author={Fischer, Michael and Georgiev, Iliyan and Groueix, Thibault and Kim, Vladimir G and Ritschel, Tobias and Deschaintre, Valentin},
  journal={arXiv preprint arXiv:2411.19322},
  year={2024}
}

@inproceedings{chajdas2010assisted,
  title={Assisted texture assignment},
  author={Chajdas, Matth{\"a}us G and Lefebvre, Sylvain and Stamminger, Marc},
  booktitle={Proceedings of the 2010 ACM SIGGRAPH symposium on Interactive 3D Graphics and Games},
  pages={173--179},
  year={2010}
}

@article{sharma2023materialistic,
  title={Materialistic: Selecting similar materials in images},
  author={Sharma, Prafull and Philip, Julien and Gharbi, Micha{\"e}l and Freeman, Bill and Durand, Fredo and Deschaintre, Valentin},
  journal={ACM Transactions on Graphics},
  volume={42},
  number={4},
  year={2023},
  publisher={ACM}
}

@inproceedings{huang2025material,
  title={Material anything: Generating materials for any 3d object via diffusion},
  author={Huang, Xin and Wang, Tengfei and Liu, Ziwei and Wang, Qing},
  booktitle=CVPR,
  pages={26556--26565},
  year={2025}
}

@article{zhang2024dreammat,
  title={Dreammat: High-quality pbr material generation with geometry-and light-aware diffusion models},
  author={Zhang, Yuqing and Liu, Yuan and Xie, Zhiyu and Yang, Lei and Liu, Zhongyuan and Yang, Mengzhou and Zhang, Runze and Kou, Qilong and Lin, Cheng and Wang, Wenping and others},
  journal={ACM Transactions on Graphics (TOG)},
  volume={43},
  number={4},
  pages={1--18},
  year={2024},
  publisher={ACM New York, NY, USA}
}

@article{secord2011perceptual,
  title={Perceptual models of viewpoint preference},
  author={Secord, Adrian and Lu, Jingwan and Finkelstein, Adam and Singh, Manish and Nealen, Andrew},
  journal={ACM Transactions on Graphics (TOG)},
  volume={30},
  number={5},
  pages={1--12},
  year={2011},
  publisher={ACM New York, NY, USA}
}

@article{khosla2020supervised,
  title={Supervised contrastive learning},
  author={Khosla, Prannay and Teterwak, Piotr and Wang, Chen and Sarna, Aaron and Tian, Yonglong and Isola, Phillip and Maschinot, Aaron and Liu, Ce and Krishnan, Dilip},
  journal=NIPS,
  volume={33},
  pages={18661--18673},
  year={2020}
}

@inproceedings{wang2015sketch,
  title={Sketch-based 3d shape retrieval using convolutional neural networks},
  author={Wang, Fang and Kang, Le and Li, Yi},
  booktitle=CVPR,
  pages={1875--1883},
  year={2015}
}

@inproceedings{wu2024generalizing,
  title={Generalizing single-view 3d shape retrieval to occlusions and unseen objects},
  author={Wu, Qirui and Ritchie, Daniel and Savva, Manolis and Chang, Angel X},
  booktitle={2024 International Conference on 3D Vision (3DV)},
  pages={893--902},
  year={2024},
  organization={IEEE}
}

@inproceedings{ruan2024tricolo,
  title={Tricolo: Trimodal contrastive loss for text to shape retrieval},
  author={Ruan, Yue and Lee, Han-Hung and Zhang, Yiming and Zhang, Ke and Chang, Angel X},
  booktitle=WACV,
  pages={5815--5825},
  year={2024}
}

@inproceedings{ankerst19993d,
  title={3D shape histograms for similarity search and classification in spatial databases},
  author={Ankerst, Mihael and Kastenm{\"u}ller, Gabi and Kriegel, Hans-Peter and Seidl, Thomas},
  booktitle={International symposium on spatial databases},
  pages={207--226},
  year={1999},
  organization={Springer}
}

@misc{Thea,
  author = {Chaudhuri, Siddhartha},
  title = {Thea},
  howpublished = {\url{https://github.com/sidch/Thea}},
  note={accessed Nov 2025}
  }

@article{zhang2025texverse,
  title={Texverse: A universe of 3d objects with high-resolution textures},
  author={Zhang, Yibo and Zhang, Li and Ma, Rui and Cao, Nan},
  journal={arXiv preprint arXiv:2508.10868},
  year={2025}
}

@inproceedings{chen2003visual,
  title={On visual similarity based 3D model retrieval},
  author={Chen, Ding-Yun and Tian, Xiao-Pei and Shen, Yu-Te and Ouhyoung, Ming},
  booktitle={Computer graphics forum},
  volume={22},
  number={3},
  pages={223--232},
  year={2003},
  organization={Wiley Online Library}
}

@inproceedings{hilaga2001topology,
  title={Topology matching for fully automatic similarity estimation of 3D shapes},
  author={Hilaga, Masaki and Shinagawa, Yoshihisa and Kohmura, Taku and Kunii, Tosiyasu L},
  booktitle={Proceedings of the 28th annual conference on Computer graphics and interactive techniques},
  pages={203--212},
  year={2001}
}

@inproceedings{xie2015deepshape,
  title={Deepshape: Deep learned shape descriptor for 3d shape matching and retrieval},
  author={Xie, Jin and Fang, Yi and Zhu, Fan and Wong, Edward},
  booktitle=CVPR,
  pages={1275--1283},
  year={2015}
}

@article{podolak2006planar,
  title={A planar-reflective symmetry transform for 3D shapes},
  author={Podolak, Joshua and Shilane, Philip and Golovinskiy, Aleksey and Rusinkiewicz, Szymon and Funkhouser, Thomas},
  journal = {ACM Trans. Graph.},
  pages={549--559},
  year={2006}
}

@article{mitra2006partial,
  title={Partial and approximate symmetry detection for 3d geometry},
  author={Mitra, Niloy J and Guibas, Leonidas J and Pauly, Mark},
  journal={ACM Transactions on Graphics (ToG)},
  volume={25},
  number={3},
  pages={560--568},
  year={2006},
  publisher={ACM New York, NY, USA}
}

@inproceedings{mitra2013symmetry,
  title={Symmetry in 3d geometry: Extraction and applications},
  author={Mitra, Niloy J and Pauly, Mark and Wand, Michael and Ceylan, Duygu},
  booktitle={Computer graphics forum},
  volume={32},
  number={6},
  pages={1--23},
  year={2013},
  organization={Wiley Online Library}
}

@article{lipman2010symmetry,
  title={Symmetry factored embedding and distance},
  author={Lipman, Yaron and Chen, Xiaobai and Daubechies, Ingrid and Funkhouser, Thomas},
  journal = {ACM Trans. Graph.},
  pages={1--12},
  year={2010}
}

@article{pauly2008discovering,
  title={Discovering structural regularity in 3D geometry},
  author={Pauly, Mark and Mitra, Niloy J and Wallner, Johannes and Pottmann, Helmut and Guibas, Leonidas J},
  journal = {ACM Trans. Graph.},
  pages={1--11},
  year={2008}
}

@article{huang2014near,
  title={Near-regular structure discovery using linear programming},
  author={Huang, Qixing and Guibas, Leonidas J and Mitra, Niloy J},
  journal={ACM Transactions on Graphics (TOG)},
  volume={33},
  number={3},
  pages={1--17},
  year={2014},
  publisher={ACM New York, NY, USA}
}

@inproceedings{fang20153d,
  title={3d deep shape descriptor},
  author={Fang, Yi and Xie, Jin and Dai, Guoxian and Wang, Meng and Zhu, Fan and Xu, Tiantian and Wong, Edward},
  booktitle={Proceedings of the IEEE conference on computer vision and pattern recognition},
  pages={2319--2328},
  year={2015}
}

@inproceedings{kirillov2023segment,
  title={Segment anything},
  author={Kirillov, Alexander and Mintun, Eric and Ravi, Nikhila and Mao, Hanzi and Rolland, Chloe and Gustafson, Laura and Xiao, Tete and Whitehead, Spencer and Berg, Alexander C and Lo, Wan-Yen and Dollar, Piotr and Girshick, Ross},
  booktitle={Proceedings of the IEEE/CVF international conference on computer vision},
  pages={4015--4026},
  year={2023}
}

@inproceedings{ravisam,
  title={SAM 2: Segment Anything in Images and Videos},
  author={Ravi, Nikhila and Gabeur, Valentin and Hu, Yuan-Ting and Hu, Ronghang and Ryali, Chaitanya and Ma, Tengyu and Khedr, Haitham and R{\"a}dle, Roman and Rolland, Chloe and Gustafson, Laura and others},
  booktitle={The Thirteenth International Conference on Learning Representations},
  year={2025}
}

@article{yi2016scalable,
  title={A scalable active framework for region annotation in 3d shape collections},
  author={Yi, Li and Kim, Vladimir G and Ceylan, Duygu and Shen, I-Chao and Yan, Mengyan and Su, Hao and Lu, Cewu and Huang, Qixing and Sheffer, Alla and Guibas, Leonidas},
  journal={ACM Transactions on Graphics (ToG)},
  volume={35},
  number={6},
  pages={1--12},
  year={2016},
  publisher={ACM New York, NY, USA}
}

@inproceedings{li2022grounded,
  title={Grounded language-image pre-training},
  author={Li, Liunian Harold and Zhang, Pengchuan and Zhang, Haotian and Yang, Jianwei and Li, Chunyuan and Zhong, Yiwu and Wang, Lijuan and Yuan, Lu and Zhang, Lei and Hwang, Jenq-Neng and others},
  booktitle={Proceedings of the IEEE/CVF conference on computer vision and pattern recognition},
  pages={10965--10975},
  year={2022}
}

@inproceedings{abdelreheem2023satr,
  title={Satr: Zero-shot semantic segmentation of 3d shapes},
  author={Abdelreheem, Ahmed and Skorokhodov, Ivan and Ovsjanikov, Maks and Wonka, Peter},
  booktitle={Proceedings of the IEEE/CVF International Conference on Computer Vision},
  pages={15166--15179},
  year={2023}
}

@inproceedings{kim2024partstad,
  title={Partstad: 2d-to-3d part segmentation task adaptation},
  author={Kim, Hyunjin and Sung, Minhyuk},
  booktitle={European Conference on Computer Vision},
  pages={422--439},
  year={2024},
  organization={Springer}
}

@inproceedings{liu2023partslip,
  title={Partslip: Low-shot part segmentation for 3d point clouds via pretrained image-language models},
  author={Liu, Minghua and Zhu, Yinhao and Cai, Hong and Han, Shizhong and Ling, Zhan and Porikli, Fatih and Su, Hao},
  booktitle={Proceedings of the IEEE/CVF conference on computer vision and pattern recognition},
  pages={21736--21746},
  year={2023}
}

@inproceedings{abdelreheem2023zero,
  title={Zero-shot 3d shape correspondence},
  author={Abdelreheem, Ahmed and Eldesokey, Abdelrahman and Ovsjanikov, Maks and Wonka, Peter},
  booktitle={SIGGRAPH Asia 2023 Conference Papers},
  pages={1--11},
  year={2023}
}

@inproceedings{xue2025zerops,
  title={Zerops: High-quality cross-modal knowledge transfer for zero-shot 3d part segmentation},
  author={Xue, Yuheng and Chen, Nenglun and Liu, Jun and Sun, Wenyun},
  booktitle={2025 International Conference on 3D Vision (3DV)},
  pages={1328--1339},
  year={2025},
  organization={IEEE}
}

@article{zhou2023partslippp,
  title={Partslip++: Enhancing low-shot 3d part segmentation via multi-view instance segmentation and maximum likelihood estimation},
  author={Zhou, Yuchen and Gu, Jiayuan and Li, Xuanlin and Liu, Minghua and Fang, Yunhao and Su, Hao},
  journal={arXiv preprint arXiv:2312.03015},
  year={2023}
}

@inproceedings{ma2025find,
  title={Find any part in 3d},
  author={Ma, Ziqi and Yue, Yisong and Gkioxari, Georgia},
  booktitle={Proceedings of the IEEE/CVF International Conference on Computer Vision},
  pages={7818--7827},
  year={2025}
}

@inproceedings{umam2024partdistill,
  title={Partdistill: 3d shape part segmentation by vision-language model distillation},
  author={Umam, Ardian and Yang, Cheng-Kun and Chen, Min-Hung and Chuang, Jen-Hui and Lin, Yen-Yu},
  booktitle={Proceedings of the IEEE/CVF Conference on Computer Vision and Pattern Recognition},
  pages={3470--3479},
  year={2024}
}

@article{tang2024segment,
  title={Segment any mesh: Zero-shot mesh part segmentation via lifting segment anything 2 to 3d},
  author={Tang, George and Zhao, William and Ford, Logan and Benhaim, David and Zhang, Paul},
  journal={arXiv e-prints},
  pages={arXiv--2408},
  year={2024}
}

@inproceedings{kim2024garfield,
  title={Garfield: Group anything with radiance fields},
  author={Kim, Chung Min and Wu, Mingxuan and Kerr, Justin and Goldberg, Ken and Tancik, Matthew and Kanazawa, Angjoo},
  booktitle={Proceedings of the IEEE/CVF Conference on Computer Vision and Pattern Recognition},
  pages={21530--21539},
  year={2024}
}

@inproceedings{ying2024omniseg3d,
  title={Omniseg3d: Omniversal 3d segmentation via hierarchical contrastive learning},
  author={Ying, Haiyang and Yin, Yixuan and Zhang, Jinzhi and Wang, Fan and Yu, Tao and Huang, Ruqi and Fang, Lu},
  booktitle={Proceedings of the IEEE/CVF Conference on Computer Vision and Pattern Recognition},
  pages={20612--20622},
  year={2024}
}

@inproceedings{he2024view,
  title={View-Consistent Hierarchical 3D Segmentation Using Ultrametric Feature Fields},
  author={He, Haodi and Stearns, Colton and Harley, Adam W and Guibas, Leonidas J},
  booktitle={European Conference on Computer Vision},
  pages={268--286},
  year={2024},
  organization={Springer}
}

@article{liu2023openshape,
  title={Openshape: Scaling up 3d shape representation towards open-world understanding},
  author={Liu, Minghua and Shi, Ruoxi and Kuang, Kaiming and Zhu, Yinhao and Li, Xuanlin and Han, Shizhong and Cai, Hong and Porikli, Fatih and Su, Hao},
  journal={Advances in neural information processing systems},
  volume={36},
  pages={44860--44879},
  year={2023}
}

@inproceedings{wang2025describe,
  title={Describe, Adapt and Combine: Empowering CLIP Encoders for Open-set 3D Object Retrieval},
  author={Wang, Zhichuan and Zhou, Yang and Liu, Zhe and Yu, Rui and Bai, Song and Wang, Yulong and He, Xinwei and Bai, Xiang},
  booktitle={Proceedings of the IEEE/CVF International Conference on Computer Vision},
  pages={21026--21036},
  year={2025}
}

@article{guerrero2025fine,
  title={Fine-Grained Spatially Varying Material Selection in Images},
  author={Guerrero-Viu, Julia and Fischer, Michael and Georgiev, Iliyan and Garces, Elena and Gutierrez, Diego and Masia, Belen and Deschaintre, Valentin},
  journal = {ACM Trans. Graph.},
  year={2025}
}

@inproceedings{nair2010rectified,
  title={Rectified linear units improve restricted boltzmann machines},
  author={Nair, Vinod and Hinton, Geoffrey E},
  booktitle=ICML,
  pages={807--814},
  year={2010}
}

@article{kingma2014adam,
  title={Adam: A method for stochastic optimization},
  author={Kingma, Diederik P and Ba, Jimmy},
  journal={ICLR},
  year={2015}
}

@article{oquabdinov2,
  title={DINOv2: Learning Robust Visual Features without Supervision},
  author={Oquab, Maxime and Darcet, Timoth{\'e}e and Moutakanni, Th{\'e}o and Vo, Huy V and Szafraniec, Marc and Khalidov, Vasil and Fernandez, Pierre and HAZIZA, Daniel and Massa, Francisco and El-Nouby, Alaaeldin and others},
  journal={Transactions on Machine Learning Research},
  year={2024}
}

@article{simeoni2025dinov3,
  title={Dinov3},
  author={Sim{\'e}oni, Oriane and Vo, Huy V and Seitzer, Maximilian and Baldassarre, Federico and Oquab, Maxime and Jose, Cijo and Khalidov, Vasil and Szafraniec, Marc and Yi, Seungeun and Ramamonjisoa, Micha{\"e}l and others},
  journal={arXiv preprint arXiv:2508.10104},
  year={2025}
}

@inproceedings{zhai2023sigmoid,
  title={Sigmoid loss for language image pre-training},
  author={Zhai, Xiaohua and Mustafa, Basil and Kolesnikov, Alexander and Beyer, Lucas},
  booktitle={Proceedings of the IEEE/CVF international conference on computer vision},
  pages={11975--11986},
  year={2023}
}

@inproceedings{objaverse,
  title={Objaverse: A universe of annotated 3d objects},
  author={Deitke, Matt and Schwenk, Dustin and Salvador, Jordi and Weihs, Luca and Michel, Oscar and VanderBilt, Eli and Schmidt, Ludwig and Ehsani, Kiana and Kembhavi, Aniruddha and Farhadi, Ali},
  booktitle={Proceedings of the IEEE/CVF Conference on Computer Vision and Pattern Recognition},
  pages={13142--13153},
  year={2023}
}

@article{wang2018learning,
  title={Learning to group and label fine-grained shape components},
  author={Wang, Xiaogang and Zhou, Bin and Fang, Haiyue and Chen, Xiaowu and Zhao, Qinping and Xu, Kai},
  journal={ACM Transactions on Graphics (TOG)},
  volume={37},
  number={6},
  pages={1--14},
  year={2018},
  publisher={ACM New York, NY, USA}
}

@inproceedings{jones2022neurally,
  title={The neurally-guided shape parser: Grammar-based labeling of 3d shape regions with approximate inference},
  author={Jones, R Kenny and Habib, Aalia and Hanocka, Rana and Ritchie, Daniel},
  booktitle={Proceedings of the IEEE/CVF Conference on Computer Vision and Pattern Recognition},
  pages={11614--11623},
  year={2022}
}
}

\clearpage

\appendix
\renewcommand{\thesection}{\Alph{section}}

\twocolumn[
\centering
\vspace{-0.8em}
{\LARGE\bfseries Appendix\par}
\vspace{2em}
]

\section{Part Segmentation}
\begin{figure*}
  \centering
   \includegraphics[width=\linewidth]{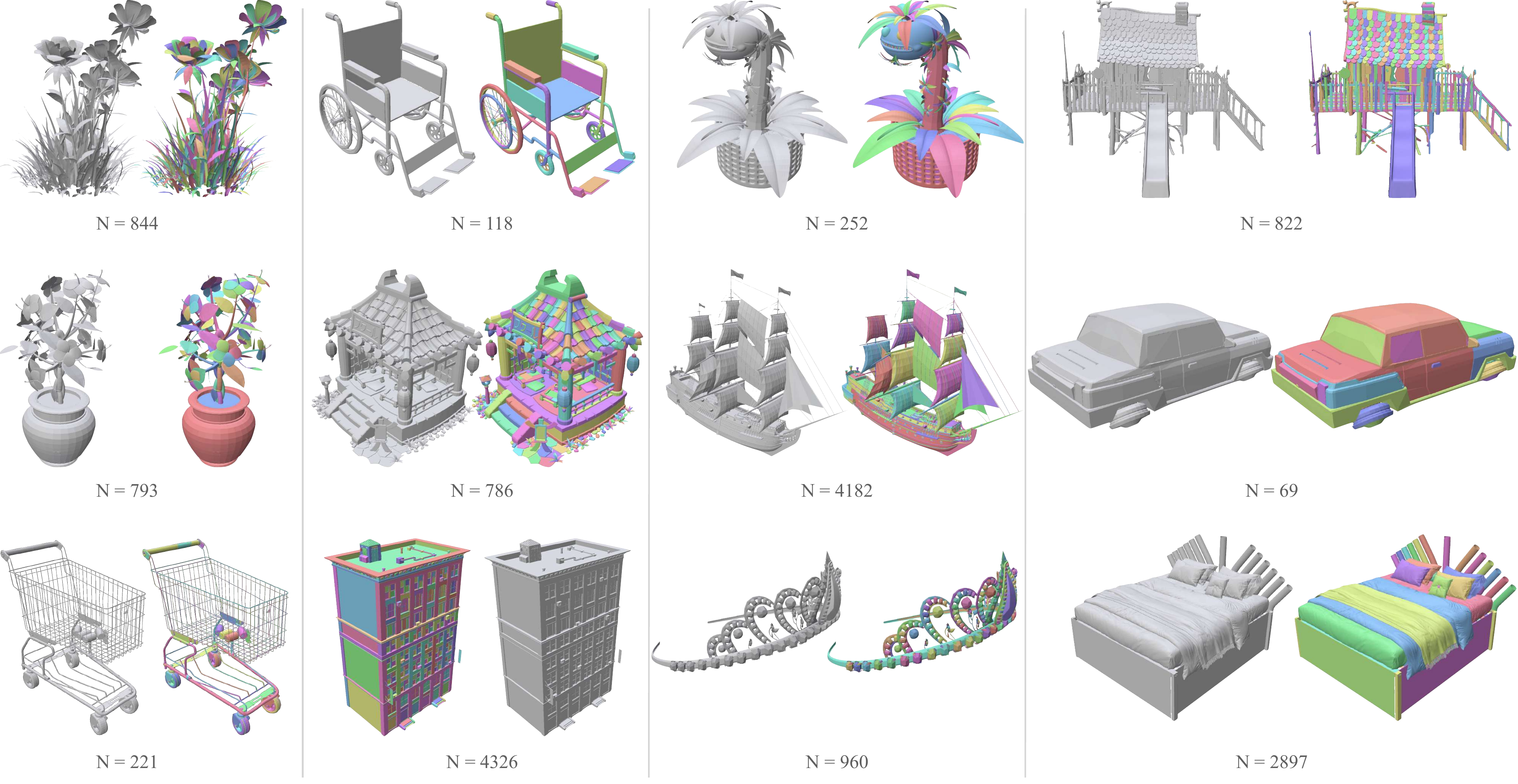}

   \caption{\textbf{Connected components as segments.} Examples of part segmentation produced using connected components. Each color corresponds to a different part. 
$N$ denotes the number of parts in the mesh, illustrating the fine granularity of the decomposition. }
   \label{fig:segments}
\end{figure*}

We obtain parts by computing connected components of each mesh after merging duplicated vertices (vertices having the exact same coordinates).
This segmentation choice reflects how artist-created meshes naturally decompose into fine-grained, meaningful components.
As shown in Figure~\ref{fig:segments}, most segments exhibit clear structural or semantic meaning (e.g., individual wheel spikes, boat ropes, bedsheet fringes).
We exclude scanned objects from our dataset since they do not exhibit such decomposition.
However, our method could be applied to scanned objects given appropriate part segmentation.
All baselines reported in the paper use the same set of parts.

\begin{figure}
  \centering
   \includegraphics[width=\linewidth]{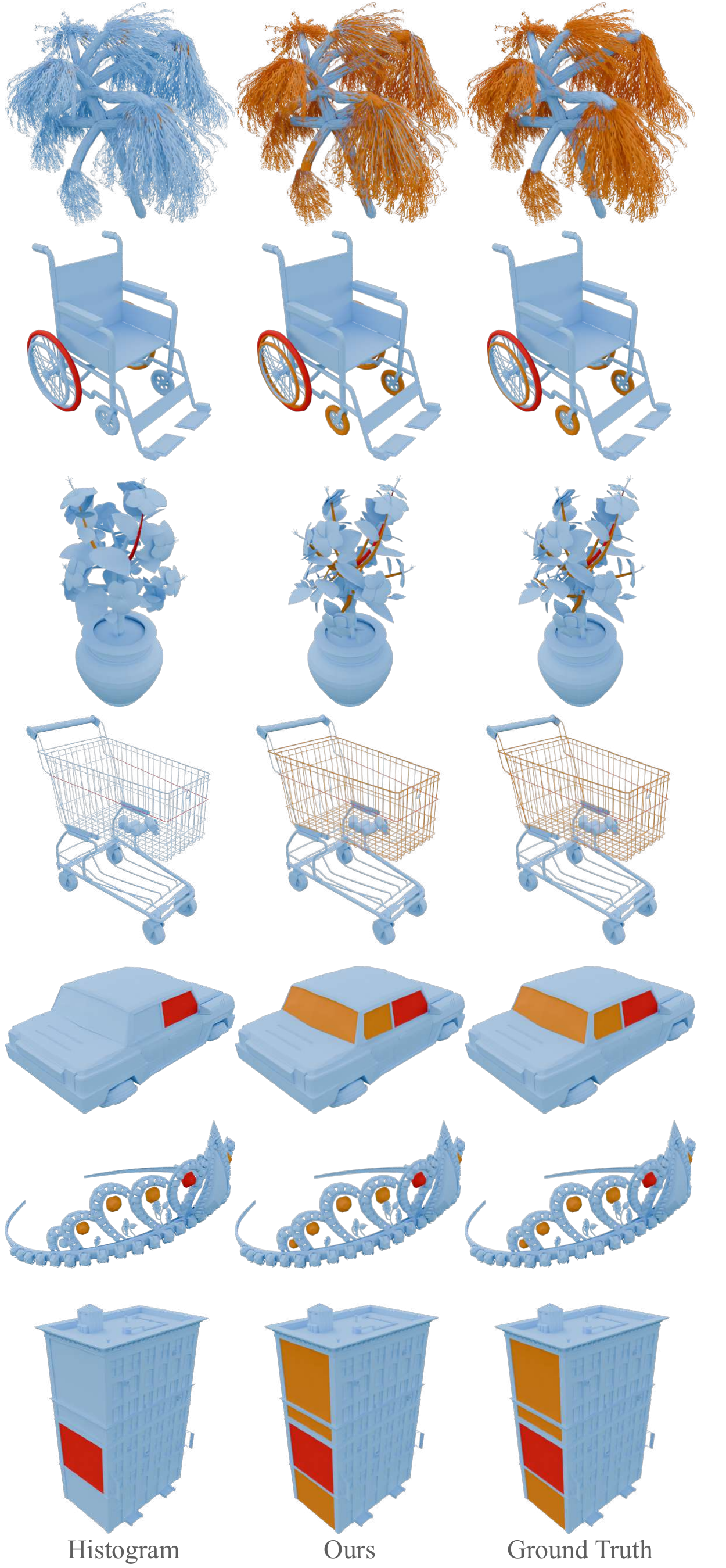}

   \caption{\textbf{Qualitative results of Histogram Matching approach.} Histogram matching successfully retrieves near-identical components (e.g., duplicated branches or similar jewels) but fails to capture structurally similar yet non-identical parts, such as the rear car window or varied tree leaves.}
   \label{fig:hist}
\end{figure}

\begin{figure}
  \centering
   \includegraphics[width=\linewidth]{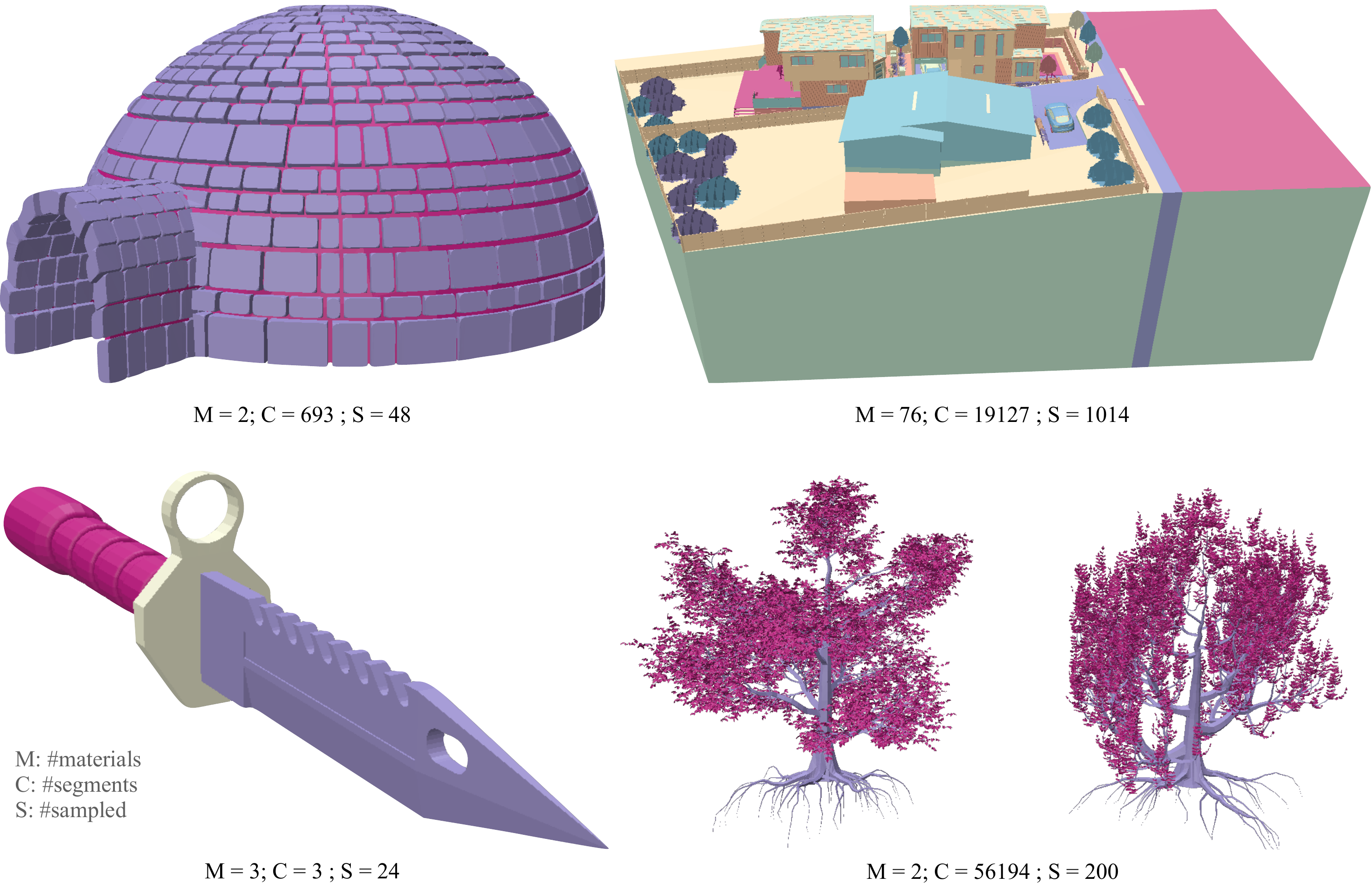}

   \caption{\textbf{Data sampling.} (M: \#materials, 
C: \#segments, S: \#sampled after balancing). \textbf{Top-left (a):} Within-mesh 
imbalance where one material (purple bricks) dominates. \textbf{Top-right (b):} 
Long-tail distribution with 76 materials but most appearing in few parts. 
\textbf{Bottom-left (c):} Insufficient samples requiring multi-view augmentation. \textbf{Bottom-right (d):} Mesh with 56K 
several near-duplicate parts requiring capping. }
   \label{fig:traindata}
\end{figure}

\section{Data balancing}~\label{sec:balance}
Objaverse presents several challenges when used directly for supervision (Table Ablation results in the main paper).
First, many meshes exhibit strong class imbalance, where a few material identities account for the majority of parts within a shape. 
Figure~\ref{fig:traindata}(a) shows such a case, where most parts share a single material label.
Second, meshes with a large number of material identities often display a pronounced long-tail distribution, with part counts per material following a power-law decay. 
As shown in Figure~\ref{fig:traindata}(b), although the mesh contains 76 materials, most appear only in a few parts.
Third, meshes with very large numbers of parts can dominate training if sampled naively. For example, Figure~\ref{fig:traindata}(d) shows a mesh with 56,194 parts, many of which are near-duplicate geometries. Such meshes inflate dataset size without contributing proportional diversity and can lead to dataset-level bias if each part is sampled uniformly.

To address geometric redundancy, we merge highly similar parts using a vertex-distribution histogram matching procedure (section~\ref{sec:hist}), yielding a more compact and diverse part set. 
We further apply data balancing strategies to mitigate both within and across mesh imbalance, resulting in a more uniform distribution over both meshes and material identities. To do so, we apply the following constraints while sampling:

\begin{itemize}
    \item Each material identity within a mesh should have at least 8 training samples. If the number of deduplicated parts for a material is below 8, we first include additional instances of the same geometry from different locations in the mesh, as these occur under different spatial context and serve as natural augmentation.
If this is still insufficient (e.g., Fig~\ref{fig:traindata}(c)), we render multiple additional viewpoints of the same part, treating each viewpoint as an additional training sample.
\item We limit each material from each mesh to a maximum of 100 samples across the training set to avoid overrepresentation of highly frequent materials. 
    \item For each mesh, we constrain the ratio between the most frequent and least frequent material (in terms of part count) to be at most 5, preventing single dominant materials from overwhelming supervision.
    
\end{itemize}

We choose these hyperparameters heuristically to balance training scale with data diversity.
After balancing, the final training set contains 1.9M part samples drawn from 22k meshes, and data generation takes approximately two days on 4 NVIDIA L40 GPUs.

\section{Histogram matching}\label{sec:hist}

We use a geometry based approach~\cite{ankerst19993d,Thea} to identify the near-duplicate parts within each mesh.
For each part, we compute a histogram of vertex distances measured from the part's centroid, forming a fixed-length descriptor of its overall vertex distribution.
Two parts are considered duplicates only if (i) their vertex counts differ by less than $5\%$, (ii) their histograms fall within a small $\ell_1$ difference threshold, and (iii) their overall scales (maximum radial extent) agree within a relative tolerance. 
We use conservative thresholds (64 histogram bins, $\ell_1$ threshold $10^{-2}$, scale tolerance $10^{-2}$) to minimize false positives.
From each duplicate group, we retain a single representative part, except in cases where additional instances are intentionally reused for data augmentation as described in Section~\ref{sec:balance}. 
We show the qualitative results from Histogram Matching baseline in Figure~\ref{fig:hist}. It successfully identifies near-identical parts but misses those with moderate geometric variation.

\begin{figure}[h]
  \centering
  \includegraphics[width=0.8\linewidth]{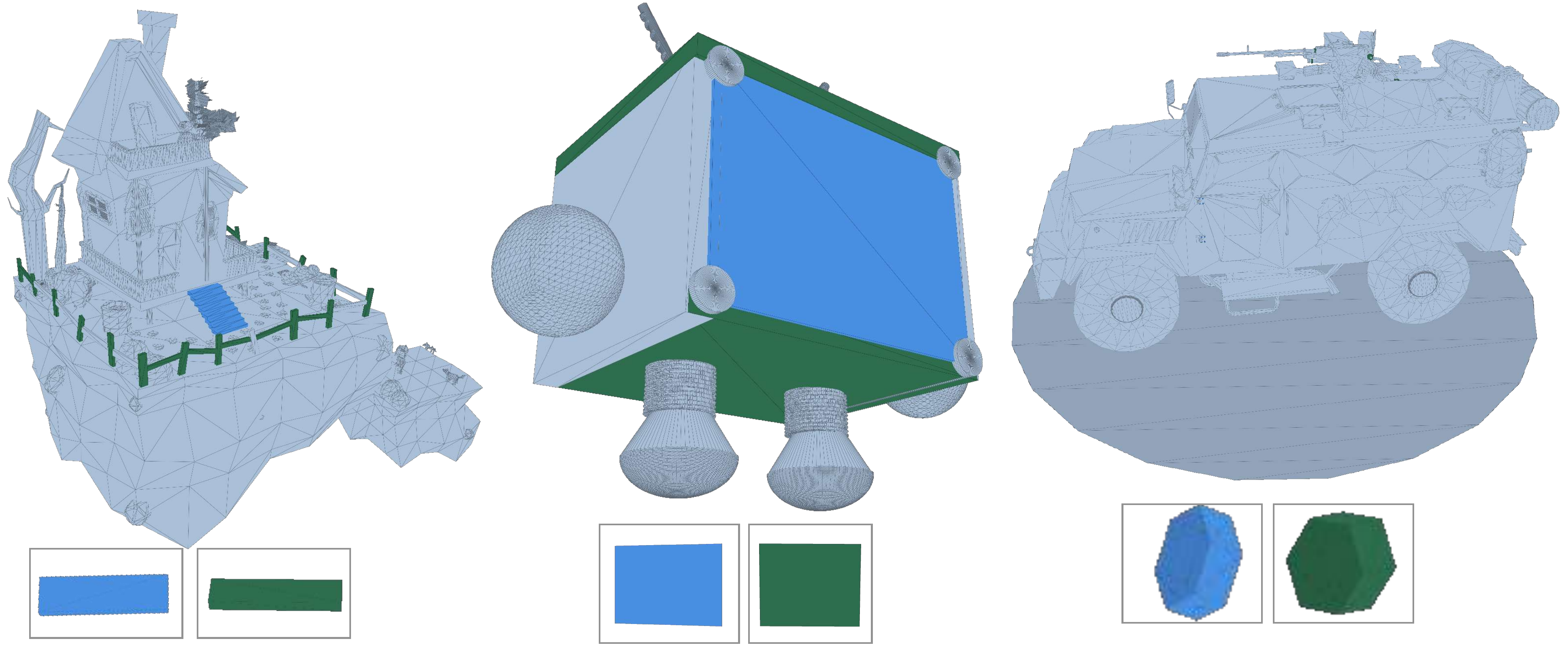}
  \caption{Deduplication error when using similar pieces across categories (e.g., cuboid for \textcolor{staicrs}{stairs} and \textcolor{fencce}{fence}).}
  \label{fig:rebuttal}
\end{figure}

This deduplication step and learned embeddings serve different roles: histogram matching only collapses near-identical rigid-transformed duplicates to shrink the search space, while embeddings retrieve geometrically varied parts with shared material. 
We apply deduplication during training to reduce redundancy and improve batch diversity, and at inference because rendering and encoding dominates the runtime, as discussed in section~\ref{sec:runtime}. 
Table~\ref{tab:dedup_inference} shows that deduplication at inference yields a median $4.4\times$ and up to $520\times$ runtime speedup. 
Histogram matching itself accounts for only $0.16\%$ of total \textit{Dedup} time. 
The small accuracy drop without deduplication stems from increased chance of poor viewpoints; deduplication enables more stable canonical renderings. 
We use conservative thresholds, merging only essentially identical parts; a manual audit shows a $1.8\%$ false-merge rate over 2349 groups across 141 meshes (Fig.~\ref{fig:rebuttal}), typically when identical structures are used across different materials. 
Importantly, the histogram-matching baseline remains far below our method in AUC-PR, confirming deduplication does not solve grouping task.

\begin{table}
\centering
\small
\setlength{\tabcolsep}{6pt}
\caption{Effect of deduplication on inference runtime.}
\begin{tabular}{lcccc}
\hline
Inference & \#Parts\textsuperscript{$\dagger$} & Time(s)\textsuperscript{$\dagger$} & AUC $\uparrow$ & mAP $\uparrow$ \\
\hline
\textit{No Dedup}  &  265; 1144.2 & 48.91; 208.05  & 88.13 &89.68 \\
\textit{Dedup}  &  60; 181.0 &  11.14; 33.01&  89.74 & 91.70   \\
\bottomrule
\multicolumn{5}{c}{\footnotesize \textsuperscript{$\dagger$}Reported numbers are (Median; IQR) over full test set.}
\end{tabular}
\label{tab:dedup_inference}
\end{table}

\begin{table*}
    \rowcolors{2}{white}{uoftcoolgray!25}
    \centering
    \caption{Performance comparison of different DINO backbone variants on our material-aware part retrieval benchmark. Results show that scaling to larger models provides only marginal or inconsistent gains across metrics. All metrics in (\%).}
    \label{tab:dinovariants}
    \setlength{\tabcolsep}{5pt}
    \renewcommand{\arraystretch}{1.1}
    \begin{tabular}{lccccccc}
        \toprule
        \textbf{Method} & \textbf{AUC PR} & \textbf{R-Prec} & \textbf{mAP} & \multicolumn{4}{c}{\textbf{Recall@K}} \\
        \cmidrule(lr){5-8}
         &  &  &  & \textbf{R@5} & \textbf{R@10} & \textbf{R@20} & \textbf{R@100} \\
        \hspace{1em}Dino v2 (small) & 80.13 & 77.33  &82.23  & 33.39 & 45.14 & 57.03 & 81.60   \\
        \hspace{1em}Dino v2 (base) & 78.99 & 76.24  & 81.98 & 33.37 & 45.10 & 56.73 & 81.00   \\
       \hspace{1em}Dino v2 (large) & 78.28 & 74.41  & 80.59 &  33.94&  44.70&  55.92& 81.00  \\
        \hspace{1em}Dino v2 (giant) & 78.30 & 76.51  & 81.78 &  33.57&  45.15&  56.29& 80.47  \\
        \hline
         \hspace{1em}Dino v3 (small) &  \textbf{81.14} & 78.32  & 83.49 & 34.57 & 45.49 & 56.63 & 82.18  \\
        \hspace{1em}Dino v3 (base) & 80.26 &   77.07&  82.08& 33.52 & 44.65 & 55.60 & 81.22  \\
         \hspace{1em}Dino v3 (large) &  \underline{80.86}&  78.92 & 84.01 &35.18  &47.09  &57.70  & 81.83  \\
         \hspace{1em}Dino v3 (huge) &  79.50&78.03 & 83.11 & 33.76 &45.11  & 56.89 &80.98     \\
         \hspace{1em}Dino v3 (7B) &  79.27& 77.89  & 83.13 & 34.26 & 45.91 & 56.98 & 81.76  \\
        
        \bottomrule
    \end{tabular}
\end{table*}
\section{Additional results}

\subsection{Qualitative results on Objaverse}
\begin{figure}[t]
  \centering
  \includegraphics[width=\linewidth]{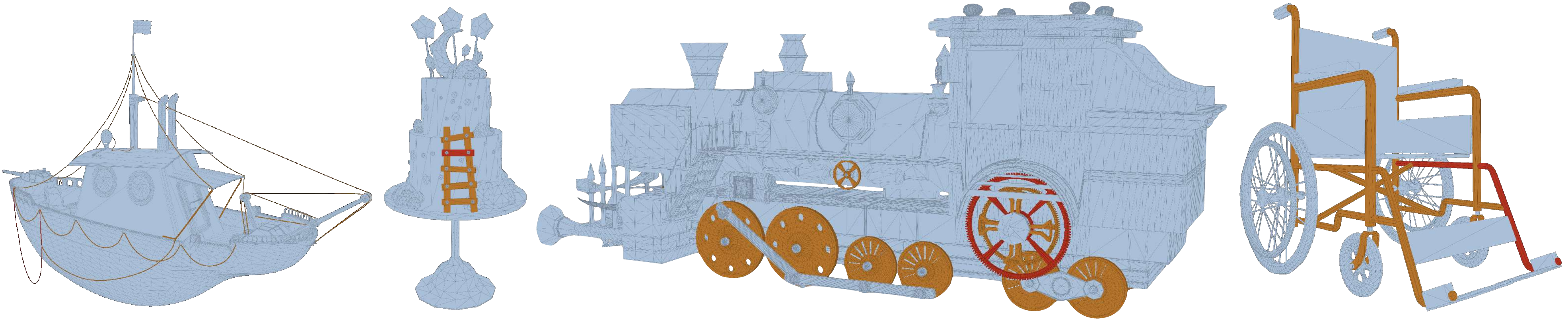}
  \caption{Additional qualitative results on Objaverse test set.}
  \label{fig:more_variety}
\end{figure}

In Figure~\ref{fig:more_variety}, we present additional results from the Objaverse test benchmark showing diverse structures that can be grouped based on material (ropes across the boat, ladder body, heterogeneous locomotive wheels, wheelchair frame), demonstrating robustness under structural differences.

\subsection{Qualitative results on TexVerse}

To evaluate generalization beyond Objaverse, we qualitatively test our method on meshes from TexVerse~\cite{zhang2025texverse}.
We select meshes that are artist-generated (not scanned), decomposable into connected components, and contain non-identical parts that share the same material, which narrows the selection pool.

\begin{figure*}
  \centering
   \includegraphics[width=\linewidth]{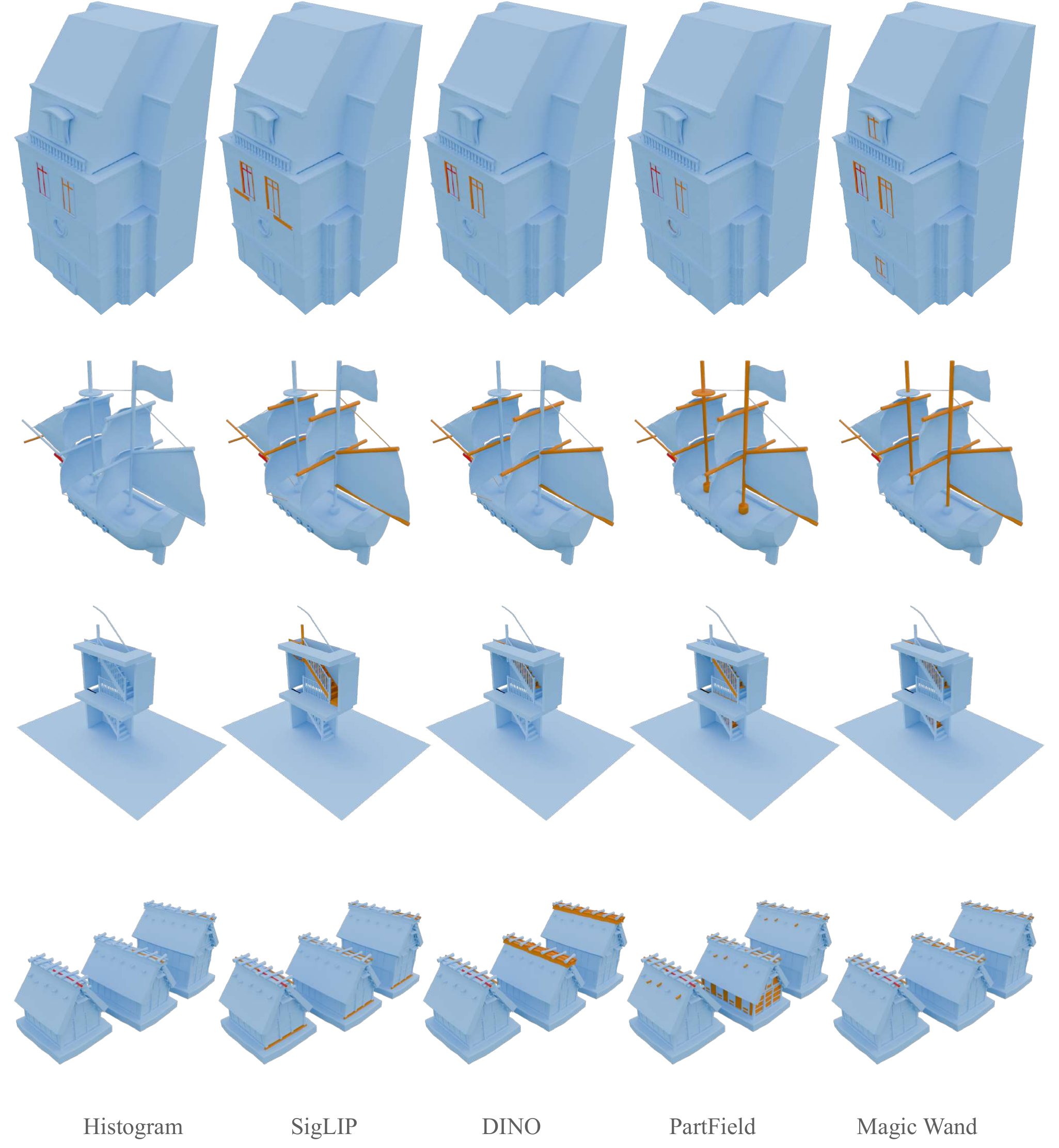}

   \caption{\textbf{Results from Texverse.} For each mesh in a row, the red part denotes the query and the orange parts indicate the retrieved matches. While baselines often miss non-identical instances or include unrelated parts, our method can retrieve material-consistent components (window frames, ship masts, stairs, and roof logs).}
   \label{fig:texverse}
\end{figure*}

Figure~\ref{fig:texverse} shows the qualitative results. 
We observe similar trends as in our test bench from Objaverse. 
Baselines often miss valid instances, such as non-identical window frames on the house, vertical masts on the boat, lower-floor stair steps, or roof logs, or retrieve unrelated structures such as bricks on the house, ropes on the boat, stair stringers, or rooftops in the huts.
In contrast, our method retrieves these material-consistent parts across geometric and structural variations more reliably.

\subsection{DINO variants}
 We finetune our supervised model from DINO-v3 \textit{small} backbone. 
To test the effect of backbone scale, we also evaluate multiple DINO-v2 and DINO-v3 variants, as summarized in Table~\ref{tab:dinovariants}. 
While larger models offer marginal improvements in some metrics (e.g., mAP), we do not observe consistent gains across all evaluation criteria. 
Given its competitive performance and fewer parameters, we adopt DINO-v3 \textit{small} in all our experiments.

\begin{figure*}
  \centering
   \includegraphics[width=\linewidth]{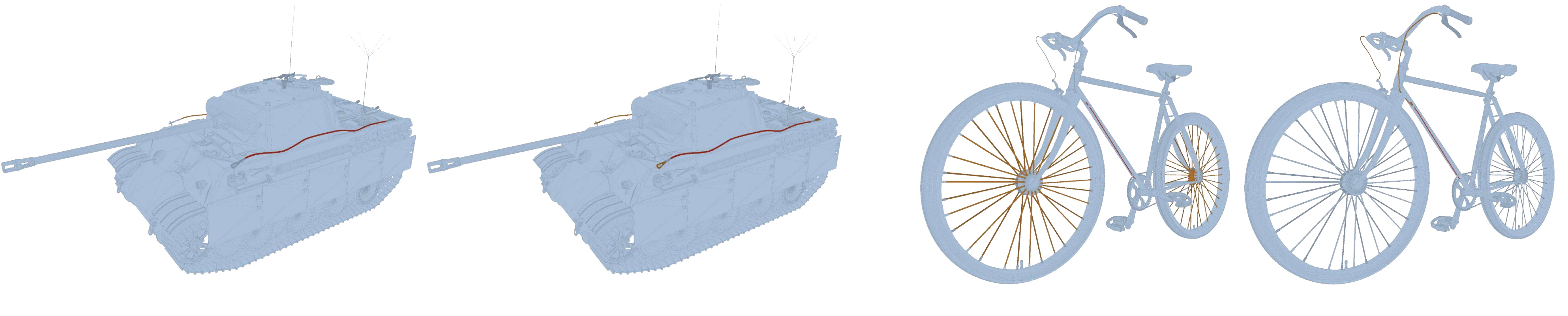}

   \caption{\textbf{Geometric variation within a material.} Given a query part (red), we show retrieved parts (left in each pair) versus ground truth (right in each pair).  \textit{Left:} Cable endpoints in the machinery exhibit  different geometry from the cable body due to folding and knotting. \textit{Right:} Bike cables near the 
   down tube are tightly coiled, appearing geometrically distinct from the loosely suspended cable segments near the handlebars. Such cases can result in incomplete retrieval.}
   \label{fig:failure1}
\end{figure*}

\begin{figure*}
  \centering
   \includegraphics[width=\linewidth]{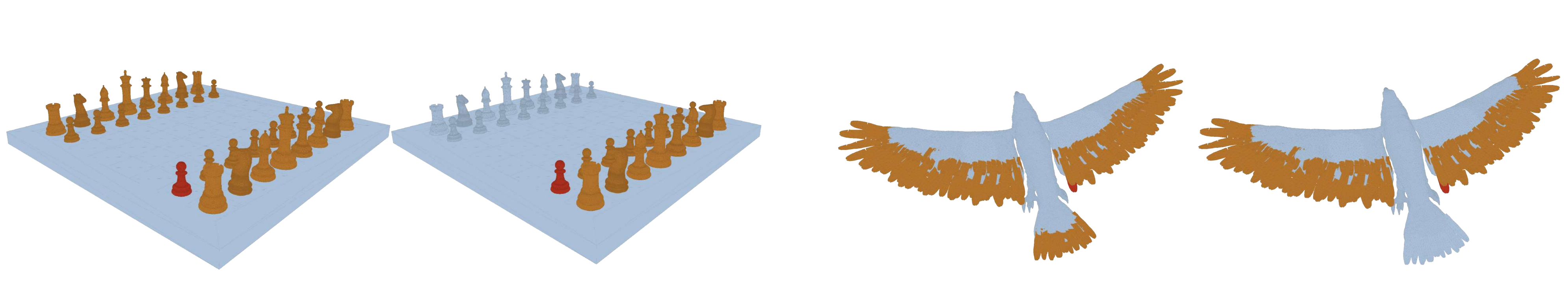}

   \caption{\textbf{Geometric similarity across different materials.} Given a query part (red), we show retrieved parts (left in each pair) versus ground truth (right in each pair). \textit{Left:} Chess pieces have identical geometry across opposing teams, differing only in material. \textit{Right:} Tail and wing feathers share similar geometric structure, leading to incorrect grouping despite different material assignments. These cases could be mitigated by training on more examples with similar contextual patterns.}
   \label{fig:failure2}
\end{figure*}

\begin{figure*}
  \centering
   \includegraphics[width=\linewidth]{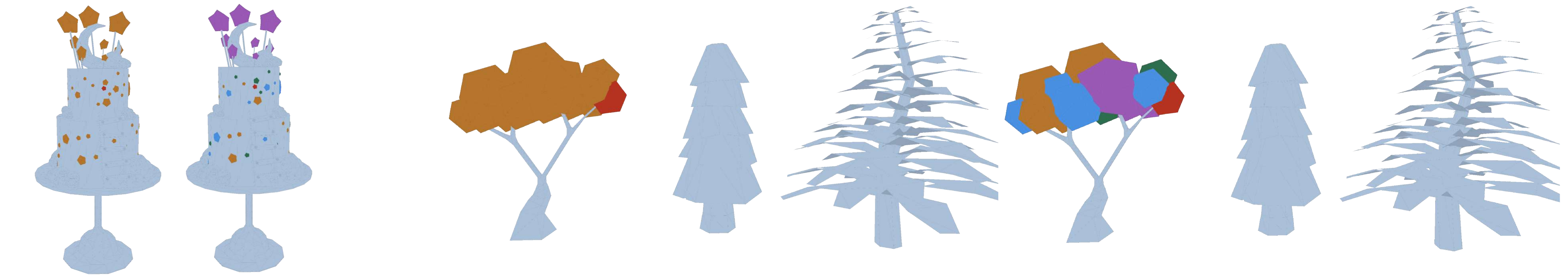}

   \caption{\textbf{Ambiguity arising from artistic intent.} Given a query part (red), we show retrieved parts (left in each pair) versus ground truth (right in each pair). \textit{Left:} Our method retrieves all pentagon-shaped cake toppings based on geometric and contextual similarity, but the actual assignments had different materials (shown in varying colors). \textit{Right:} We retrieve all foliage when querying one leaf cluster, but the mesh contains intentional material variation across the tree. Such artistic choices represent inherent ambiguity that cannot be resolved from geometry and context.}
   \label{fig:failure3}
\end{figure*}

\section{Limitation and failure cases}

Despite strong performance, our methods can face limitations:

\paragraph{Large geometric variation within the same material} Our model relies partly on geometric cues to infer material consistency. 
When parts share a material but differ significantly in shape, retrieval may be incomplete (Figure~\ref{fig:failure1}). 
For example, machinery cables that are folded and knotted at their endpoints appear geometrically distinct from their linear midsections. While our method successfully retrieves similar mid cable segments, it misses the endpoints. 

\paragraph{Geometric similarity across different materials} In some cases, when parts from different materials share highly similar geometry and local context, our method can incorrectly group them (Figure~\ref{fig:failure2}). In a chess set, pieces from opposing teams have identical forms and differ only by material or color. 
Without sufficient training examples demonstrating such material distinctions in similar geometric contexts, the model groups them together.

\paragraph{Inherent ambiguity arising from artistic intent} Certain material assignments reflect artistic or stylistic decisions that cannot be inferred from geometry or spatial context (Figure~\ref{fig:failure3}). 
Visually similar cake toppings may be intentionally assigned different materials for aesthetic variation. 
 Our method retrieves similar parts based on learned patterns, but such artistic choices represent inherent ambiguity that cannot be resolved without additional user input.

We focus on designer-created meshes with reliable part structure; evaluation on emerging AI-generated assets is future work.

\section{Runtime} \label{sec:runtime}

For each mesh, rendering and encoding is a one-time cost. The representations can be reused for retrievals on the mesh. 

\paragraph{Rendering time} Per-part rendering of $I^{part}, I^{ctx}, I^{full}$ takes 0.138 $\pm$ 0.016s. 

\paragraph{Encoding time} For meshes with $<$ 50, 50–200, and $>$ 200 parts, the median encoding times with DINO-v3 \textit{small} are 1.11s, 3.67s, and 22.78s respectively.

\paragraph{Retrieval time} For queries on meshes with $<$ 50, 50–200, and $>$ 200 parts, the retrieval times are 0.000638s, 0.001633s, and 0.010847s respectively.

\end{document}